\documentclass{article} 
\usepackage[final]{colm2026_conference}

\usepackage{microtype}
\usepackage{hyperref}
\usepackage{url}
\usepackage{comment}
\usepackage{booktabs}
\usepackage{multirow}
\usepackage{graphicx}
\usepackage{amsmath}
\usepackage{subcaption}
\usepackage[table]{xcolor}
\usepackage{wrapfig}
\usepackage{xcolor}         
\usepackage{amsmath}
\usepackage{bm}
\usepackage{colortbl}
\usepackage{enumitem}
\usepackage{lineno}
\usepackage{booktabs}  

\definecolor{tablewhite}{HTML}{EAEFEF}

\newcommand{\modelname}{\texttt{\textbf{GRADE}}}

\definecolor{darkblue}{rgb}{0, 0, 0.5}
\hypersetup{colorlinks=true, citecolor=darkblue, linkcolor=darkblue, urlcolor=darkblue}

\title{GRADE: Probing Knowledge Gaps in LLMs through \\ Gradient Subspace Dynamics}

\author{ Yujing Wang$^{1,3}$\thanks{Yujing is an incoming PhD student at King's College London, supervised by Hanqi Yan. Correspondence to Hanqi Yan (\href{mailto:someone@example.com}{hanqi.yan@kcl.ac.uk}). }\quad Yuanbang Liang$^{2}$\quad Yu-Kun Lai$^{2}$\quad Hainan Zhang$^{1}$ \quad Hanqi Yan$^{3}$ \\ $^1$Beihang University \quad $^2$Cardiff University \quad $^3$King's College London \\ \texttt{yujing.3.wang@kcl.ac.uk \quad \{LiangY32, LaiY4\}@cardiff.ac.uk \quad hanqi.yan@kcl.ac.uk}}

\begin{document}

\ifcolmsubmission
\linenumbers
\fi

\maketitle

\begin{abstract}
Detecting whether a model's internal knowledge is sufficient to correctly answer a given question is a fundamental challenge in deploying responsible LLMs. In addition to verbalising the confidence by LLM self-report, more recent methods explore the model internals, such as the hidden states of the response tokens, to capture how much knowledge is activated. We argue that such activated knowledge may not align with what the query requires, e.g., capturing the stylistic and length-related features that are uninformative for answering the query. To fill the gap, we propose \modelname{} (\textbf{G}\textbf{R}\textbf{A}dient \textbf{D}ynamics for knowl\textbf{E}dge gap detection), which quantifies the knowledge gap via the cross-layer rank ratio of the gradient to that of the corresponding hidden state subspace. This is motivated by the property of gradients as estimators of the required knowledge updates for a given target. We validate \modelname{} on six benchmarks, demonstrating its effectiveness and robustness to input perturbations. In addition, we present a case study demonstrating how the gradient chain can generate interpretable explanations of knowledge gaps in long-form answers. Code is available at \url{https://github.com/yjEugenia/llm-aware}.


\end{abstract}

\section{Introduction}
Large Language Models (LLMs) have demonstrated remarkable capabilities across diverse reasoning tasks by eliciting internal knowledge in response to various queries~\citep{cot,mathreason}. However, before taking the generated response for granted, we need to check whether the model is capable of solving the problem~\citep{white_predentropy,boundary_survey}, which is different from uncertainty quantification, as a model can be highly confident yet still be wrong. Detecting such knowledge gaps is essential to mitigate failure modes such as hallucination~\citep{hallu_0,hallu_1}. Here, a knowledge gap refers to the disparity between a model knowing and not knowing how to answer a query, which empirically manifests as the intrinsic difference between generating correct and incorrect answers. While LLM verbalization~\citep{black_words,black_askcali,black_priorijudge} offers the most direct route to eliciting self-reported knowledge gaps, like ``\textit{Yes, I'm able to}'', recent work has shifted toward exploiting intrinsic model signals. Among these, hidden states of generated answer tokens~\citep{white_probe,white_quncertainty} have emerged as a particularly informative source for measuring a model's internal knowledge.

Inspired by recent advances in mechanistic interpretability~\citep{white_latentknowledge,white_internal,internalstates}, hidden states have been used to classify correctly and incorrectly answered reasoning samples. These representations — ranging from the hidden state of the final response token~\citep{white_latentknowledge,white_internal} to a geometric Chain-of-Embedding (CoE)~\citep{CoE} over reasoning chains — consistently outperform predominant metrics such as perplexity and entropy, with further gains achieved by supervised variants~\citep{supervised,white_probe}. The success of leveraging the encoded internal representations extends beyond knowledge gap detection to applications such as fact verification~\citep{factuality}, underscoring the powerful expressivity of LLM representations more broadly.

However, hidden states capture activated knowledge broadly, including those irrelevant to the query, such as input lengths and style variations, which do not principally affect the knowledge required to answer correctly. In other words, existing hidden-states based methods can be violated by unexpected trivial perturbations in the input. 

\begin{wrapfigure}{r}{0.30\textwidth}
    \centering
        \centering
        \includegraphics[width=\linewidth,trim={10 0 0 0},clip]{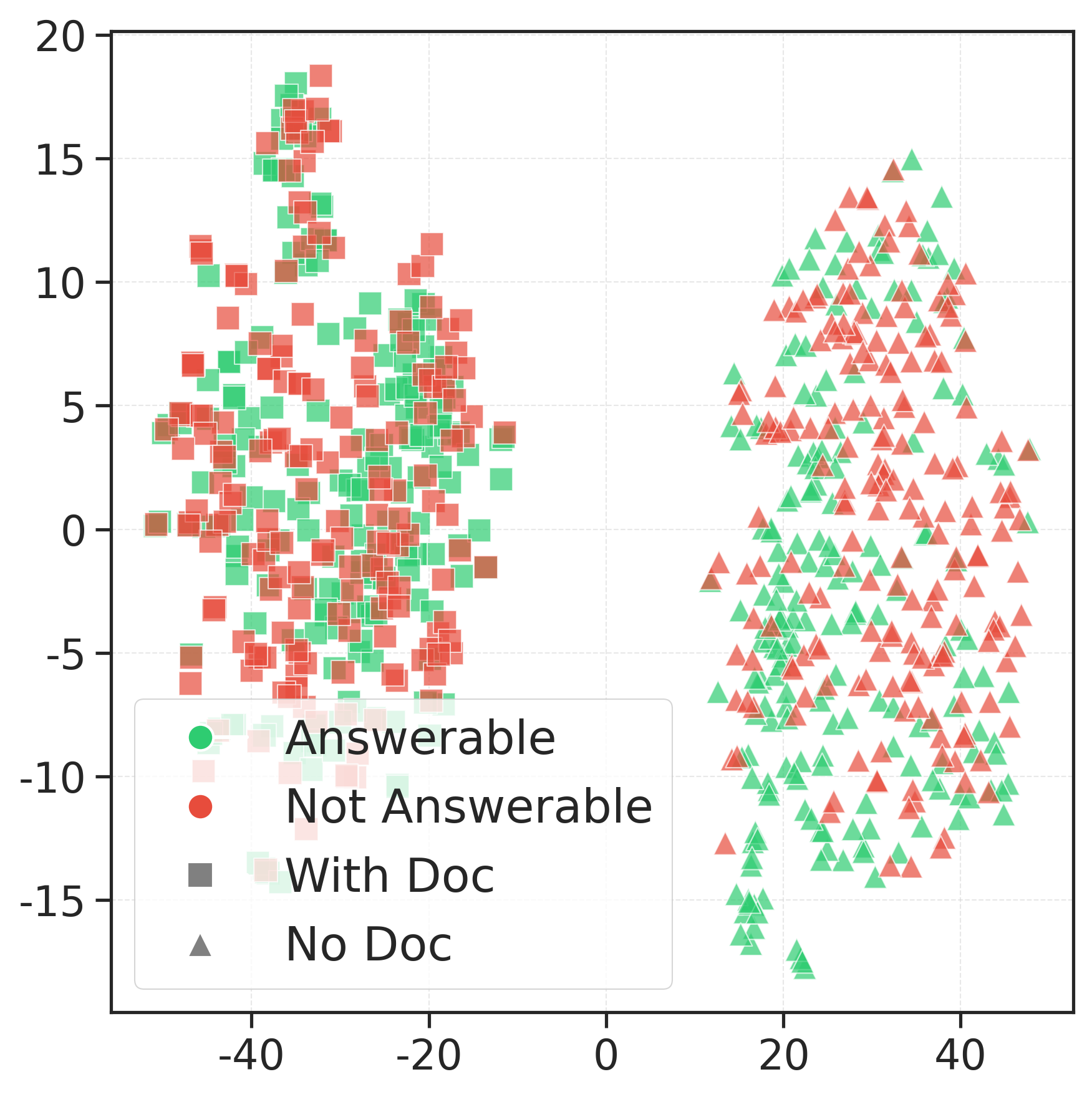} 
        \caption{t-SNE visualizations of hidden states for answerable and unanswerable queries.}
        \label{fig:tsne_hs}
    \label{fig:tsne_comparison}
\end{wrapfigure}

We illustrate this in Figure~\ref{fig:tsne_hs} using a knowledge-intensive task, where external documents can be retrieved as additional context. We prompt the LLM to generate answers both with and without retrieved documents, then project the last-layer's hidden states into 2D space via t-SNE. Red and green denote correctly and incorrectly answered responses, respectively, while squares and triangles denote responses generated with and without retrieved documents. We observe that hidden states are clearly separated by the presence or absence of retrieved documents, but \textit{\textbf{not}} by whether the answer is correct. It demonstrates that adding additional information, even when it doesn't help close the knowledge gap, can greatly affect the hidden states distribution. 

Building on these insights, we propose \modelname{}, a gradient-based dynamic analysis that measures the effective dimension of the parameter updates for the given query.
Inspired by pioneering works showing that gradients localize knowledge-sensitive parameters via gradient-based Fisher Information~\citep{fisher_1,fisher_2}, we use the rank of the gradient matrix to distinguish the knowledge structure between correctly and incorrectly answered questions.
Based on the observation that gradients lie within the subspace spanned by hidden states, we project the gradient into this shared subspace and apply a rank ratio to measure the effective dimension of query-specific update pressure represented within the activated parameter subspace. To alleviate the hyperparameter reliance, we introduce a spectrum-aware measurement that adaptively selects the most informative components for stable rank analysis, and train a probe to capture cross-layer rank ratio dynamics.

Our main contributions can be summarized below:
\begin{itemize}[itemsep=0pt,leftmargin=*]
    \item We identify a systematic problem of existing hidden state methods in knowledge gap detection, i.e., their vulnerability to irrelevant input perturbations.
    \item We propose a gradient-based dynamic analysis method \modelname{} by measuring the stable rank ratio of the projected knowledge updates, for LLM knowledge gap detection. This method also enables the interpretation generation for long-form answers.
    \item Empirically, we demonstrate that \modelname{} consistently outperforms baselines across six datasets, including question answering, multiple-choice, and math reasoning in predicting question answerability and cross-dataset transferability. 
\end{itemize}

\section{Related Work}
Knowledge gaps detection can be achieved by either merely evaluating the model's outputs (black-box) or accessing the model's internal states. \\
\textbf{Black-box.} 
\textit{Verbalization}~\citep{black_askcali,black_sayknowledge,black_priorijudge} directly prompts LLMs to self-report their confidence or acknowledge their inability to answer a given question in the generated text. Despite the further enhancements, such as self-reflection capabilities of RLHF-tuned models, \cite{black_overconf} found these methods are still largely hindered by overconfidence and miscalibration. \textit{Sampling-based methods}~\citep{black_semantic_uncertainty,black_uq_sample,black_sample_aggr,black_consistency} evaluate the semantic consistency or agreement across multiple sampled responses to the same query. For instance, semantic uncertainty \citep{black_semantic_uncertainty} estimates confidence by sampling multiple responses, clustering them by semantic equivalence, and computing entropy over the resulting clusters. However, such methods are often computationally expensive, requiring multiple sampling passes and external models for semantic consistency evaluation.\\
\textbf{White-box} methods allow access to model internals, such as generated \textit{token probability} and \textit{hidden states} (activations). Entropy \citep{white_predentropy} or perplexity \citep{white_perplexity} from the model's logits are the most-widely adopted baseline metrics for both knowledge gap~\citep{confi-boundary} or uncertainty~\citep{entropy_uncertainty}. More recent work also extends the token-level measurement to the atomic fact-level of the predictions~\citep{while_tokenlevel} and query-level~\citep{white_quncertainty}, by computing a margin over sequence token probabilities~\citep{white_tokenconsistency}.
Moving deeper into the model's architecture, i.e., internal activations, such as
MLP outputs \citep{white_internal, white_probe} and attention \citep{white_attention}. For example, \citet{white_attention} identify semantically crucial tokens by back-tracking through an attention chain to explain answer derivation. \citet{white_internal} and \citet{white_probe} train a knowledge gap detector with the hidden as inputs and factual accuracy as output label. Our \modelname{} also leverages the hidden states, but with emphasis on leveraging gradient information to alleviate the reliance on spurious features in the inputs.


\section{\modelname{} for Knowledge Gap Probing}

\begin{figure}[t]
    \centering
    \includegraphics[trim={5 5 8 5},clip, width=0.850\linewidth]{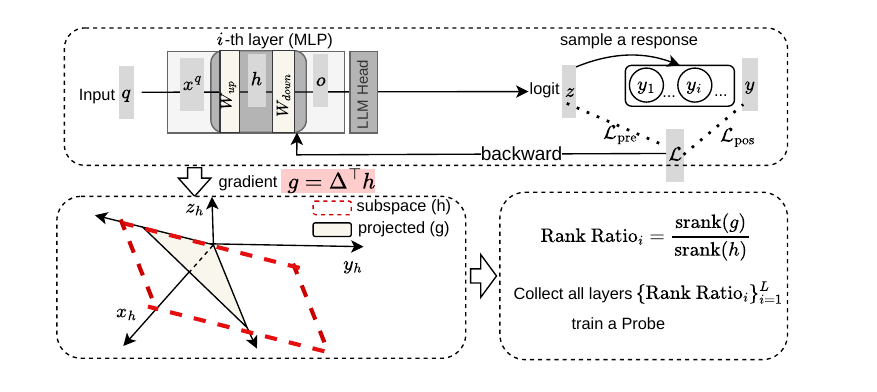}
    
\caption{ \modelname{} for knowledge gap probing. Given an input $q$, (i) \textit{Forward pass:} compute hidden states $h,o$ in the MLP block and loss $\mathcal{L}$, 
either before response generation or after; \textit{Backward pass:} derive gradient $g$; (ii) \textit{Rank ratio calculation}: project gradients onto the subspace spanned by $h$ and compute rank ratios based on stable rank (srank); (iii) \textit{Probe training:} aggregate rank ratios across $L$ layers to predict the knowledge gap.}

    \label{fig:overview}
\end{figure}

The goal of knowledge gap detection is to determine whether the model $\mathcal{M}$ possesses sufficient latent knowledge to answer a given question $q$. We formalize this task as Knowledge Boundary (KB) Detection, where the mapping function $F$ learns to map the model's internal states to a continuous score $s$ representing the knowledge gap:
\begin{equation}
F : \mathcal{I}_{\mathcal{M}}(q) \mapsto [0, 1],
\end{equation}
where $\mathcal{I}_{\mathcal{M}}(q)$ represents the updated model's internal states after feeding with $q$. Here, a knowledge gap refers to the disparity between a model knowing and not knowing how to answer a query, which empirically manifests as the intrinsic difference between generating correct and incorrect answers.

An effective detector should capture how much it contains knowledge specifically required to answer $q$.  Many hidden state-based methods~\citep{white_internal} directly use the output representations from the Multi-Layer Perceptron (MLP) blocks as the detector input, i.e., $\mathcal{I}_{M} \leftarrow MLP(q)$. However, hidden states could encode information irrelevant to the required knowledge (see Figure~\ref{fig:tsne_hs}), making them susceptible to spurious input variations. \modelname{} extracts features from $\mathcal{I}_{M}(q)$ that capture the internal computation relevant to determining knowledge sufficiency.

\paragraph{Method overview.}
We use the gradient matrix rank to distinguish the knowledge structures of correctly and incorrectly answered questions. The overview of the method is shown in Figure~\ref{fig:overview}. Given an input query $q$, a forward pass derives hidden states $h$ in the MLP blocks and loss 
$\mathcal{L}$, conditioned on whether a sampled response $y$ is available ($\mathcal{L}_{\text{pos}}$) or not ($\mathcal{L}_{\text{pre}}$). A backward pass then yields the gradients (\S\ref{subsec:info_flow}). 
To measure the proportion of generation-induced update pressure represented within the activated parameter subspace, 
we project the gradients onto the subspace spanned by $h$ and compute stable rank ratios (\S\ref{subsec:rank_proj}). Finally, rank ratios across all $L$ 
layers are aggregated as input features to train a knowledge gap detection probe (\S\ref{subsec: probe}).

\subsection{Bi-Directional Information Flow}
\label{subsec:info_flow}
\textbf{MLP feedforward.} Given a Large Language Model $\mathcal{M}$ processing a sequence of length $n$, we focus on the MLP blocks, as they are widely recognized as the primary storage for factual information \citep{mlp_knowledge}. 
Let $\bm{x^q} \in \mathbb{R}^{n \times d_{\text{model}}}$ denote the MLP input hidden states, where $d_{\text{model}}$ is the model dimension. The information flow through a gated MLP block in the LLM can be formalized as: 
\begin{align}
\label{eq_mlp}
    \bm{h} = \sigma(\bm{x^q} W_{\text{gate}}^\top ) \odot (\bm{x^q} W_{\text{up}}^\top ), \quad \bm{o} = \bm{h} W_{\text{down}}^\top.
\end{align}
During the MLP feedforward process, we first obtain the \textit{\textbf{intermediate hidden states}} $\bm{h} \in \mathbb{R}^{n \times d_{\text{ff}}}$ via the two weight matrices $W_{\text{gate}}$ and $W_{\text{up}}$, where $\sigma$ is the non-linear activation function, $\odot$ denotes the element-wise product, and $d_{\text{ff}}$ is the intermediate feed-forward dimension. The MLP output hidden states $\bm{o} \in \mathbb{R}^{n \times d_{\text{model}}}$ are then computed via the projection matrix $W_{\text{down}}$ (hereafter denoted simply as $W$).
\textbf{Generation and loss function.} From the LLM generation head, we obtain the logit $z \in \mathbb{R}^{V}$, after the projection from model dimension $d_\text{model}$ to vocabulary $V$,  
from which tokens are sampled autoregressively to form the response 
$y = [y_1, y_2, \dots, y_{\text{out}}]$. 
We then define two loss objectives $\mathcal{L}$ depending on whether the response $y$ has been generated, formally defined in Eq~(\ref{eq:loss}): 
\begin{align}
    \mathcal{L}_{pre} = -\sum_{j=1}^{V} p(z_j ) \log p(z_j), \quad \mathcal{L}_{pos} = - \sum_{t=1}^{|y_\text{out}|} \log p(y_t \mid \bm{y}_{<t},q),
\label{eq:loss}
\end{align}
where $z_j$ is the logit for the $j$-th vocabulary token, and $p(\cdot)$ denotes the softmax function.
\begin{itemize}[topsep=0pt,leftmargin=*, itemsep=0pt]
    \item For $\mathcal{L}_{pre}$, we calculate the entropy of $z$ to measure how spread out the distribution is over the vocabulary. A low entropy implies that the model knows well what to generate next given the query.
    \item For $\mathcal{L}_{pos}$, we take the model's own generated response $y$ as a pseudo-label for supervision, and then use the standard cross-entropy loss to measure how far the model's parameter updates can reach the final generation. Specifically, in long-form reasoning scenarios, the response is segmented into discrete steps based on punctuation markers. A step-wise cross-entropy loss is then applied, restricting the loss calculation to the tokens within each step. We thus generate the interpretation for the detected gap based on this mechanism, and the results are shown in \S\ref{subsec:results_explain}. 
\end{itemize}


\textbf{Backward pass.} We now apply the chain rule to derive the gradient $\bm{g}\in \mathbb{R}^{d_{\text{model}} \times d_{\text{ff}}}$ of the loss with respect to $W$. 
The gradient derivation can be expanded as:
\begin{equation}
    \bm{g} = \frac{\partial \mathcal{L}}{\partial W} = \left( \frac{\partial \mathcal{L}}{\partial \bm{o}} \right)^\top \frac{\partial \bm{o}}{\partial W} = \Delta^\top \bm{h}, 
\end{equation}
where $\Delta = \frac{\partial \mathcal{L}}{\partial \bm{o}} \in \mathbb{R}^{n \times d_{\text{model}}}$ is the intermediate signal at the pre-activation layer. Consequently, the $i$-th row of $\bm{g}$, 
$\bm{g}_i^\top \in \mathbb{R}^{d_{\text{ff}}}$, can be written as:
\begin{equation}
    \bm{g}_i^\top = \sum_{t=1}^{n} \Delta_{t,i}\, \bm{h}_t^\top,
\label{eq:subspace}
\end{equation}
a weighted sum of the input hidden states $\{\bm{h}_t\}_{t=1}^{n}$. 
Since every row of $\bm{g}$ is a linear combination of these vectors, $\bm{g}$ lies entirely within the subspace spanned by $\bm{h}$.



\paragraph{Gradient subspace illustration.} 
Based on Eq.~(\ref{eq:subspace}), the gradient lies within the subspace spanned by the hidden states, as illustrated in Figure~\ref{fig:overview}. Specifically, the gradient (white triangle) resides within the subspace (bounded by the red dashed boundary) of the 3D space ($x_h - y_h - z_h$) spanned by the hidden states. This motivates projecting both the gradient and hidden states onto their shared subspace, where the relative volume of the gradient subspace to the hidden state subspace serves as a natural proxy for the proportion of required knowledge updates within the activated knowledge. In what follows, we describe how the stable rank ratio operationalizes this proxy.


\subsection{Gradient Projection and Rank Ratio}
\label{subsec:rank_proj}

To compute the Rank Ratio as a proxy for the proportion of parameter query-specific update pressure represented within the activated parameter subspace, we first project both gradients and hidden states into a shared subspace (the region within the red dashed boundary in Figure~\ref{fig:overview}), followed by the rank ratio calculation, which also serves as an implicit normalization against input length.


\subsubsection{Projected Rank Calcualtion}
\label{subsubsec:proj}
To facilitate a direct comparison of ranks, we first project the parameter gradient into a unified representation space shared with the hidden states. Then, we employ the stable rank to robustly compute the effective dimensionality of required parameter updates.

\textbf{Projection to a shared representation space.} To measure how much gradient-introduced information occupies the total capacity of the hidden states, we take the subspace spanned by the hidden states $\bm{h}$ as the base space and project the gradient $\bm{g}$ into it. Formally, using the orthogonal projection matrix $\bm{P}_{\bm{h}} = \bm{h}^\top (\bm{h}\bm{h}^\top)^\dagger \bm{h}$ onto the row space of $\bm{h}$ \citep{inverse}, the projected gradient can be expressed as $\bm{g}\bm{P}_{\bm{h}} = \bm{A}\bm{h}$, where $\bm{A} = \bm{g}\bm{h}^\top (\bm{h}\bm{h}^\top)^\dagger$ represents the coordinate representation of the projected gradient under the hidden-state basis $\bm{h}$. 

We compute the covariance matrix of these projected gradient coordinates as $\mathcal{C}_g = \bm{A}\bm{A}^\top \in \mathbb{R}^{n \times n}$, which simplifies to:
\begin{equation}
\mathcal{C}_{g} = \mathcal{C}_h^\dagger \left( \bm{h} \bm{g}^\top \bm{g} \bm{h}^\top \right) \mathcal{C}_h^\dagger,
\end{equation}
where $\mathcal{C}_h = \bm{h}\bm{h}^\top$ is the Gram matrix of the hidden states, and $\mathcal{C}_h^\dagger$ denotes its Moore-Penrose pseudoinverse. $\mathcal{C}_g$ reflects the covariance of the gradient dynamics projected within the hidden-state space.

 
To quantify the effective dimensionality of the projected gradient space, we compute the rank of $\mathcal{C}_g$ using Singular Value Decomposition (SVD). The singular value distributions are long-tailed, and recent studies suggest that the core semantics only lie in a low-dimensional manifold, so we retain only singular values above $10^{-6}$ for rank calculation. Figure~\ref{fig:overall_rank_distribution} demonstrates the effectiveness of gradient rank 
as a discriminative signal for answerable and unanswerable questions.
 
\textbf{Stable rank for robust rank estimation.} Despite the effectiveness of naive rank, the hard threshold makes the naive rank estimate sensitive to numerical noise near the cutoff. As illustrated in Figure \ref{fig:avg_metrics}, truncating at $10^{-4}$ yields a rank of 269, whereas a naive threshold at $10^{-5}$ results in a rank of 299, leading to a substantial difference in rank calculation. 
Therefore, we employ the stable rank \citep{srank_0,srank_1} to ensure a robust and stable calculation of effective dimensionality, as it offers a smooth, spectrum-aware measure that accounts for all singular values. Formally, let $\lambda^{g}_1 \ge \lambda^{g}_2 \ge \dots \ge \lambda^{g}_n \ge 0$ be the sorted singular values of $\mathcal{C}_g$.
The stable rank applied to the two objectives:
\begin{equation}
\label{eq:srank}
    srank_{pre}(\mathcal{C}_g) = \sum_{i=1}^{n} \frac{\lambda^g_i}{\lambda^g_1},\;srank_{pos}(\mathcal{C}_g) = \sum_{i=1}^{n} \frac{(\lambda^g_i)^2}{(\lambda^g_1)^2}. 
\end{equation}
The stable rank provides a continuous measure that naturally concentrates on high-energy singular values while discounting the long tail, as illustrated in Figure \ref{fig:avg_metrics}.


\subsubsection{Rank Ratio Calculation}
\label{subsubsec: ratio}

\begin{figure}[h]
\begin{center}
    \begin{subfigure}[b]{0.3\textwidth}
        \centering
        \includegraphics[width=\textwidth]{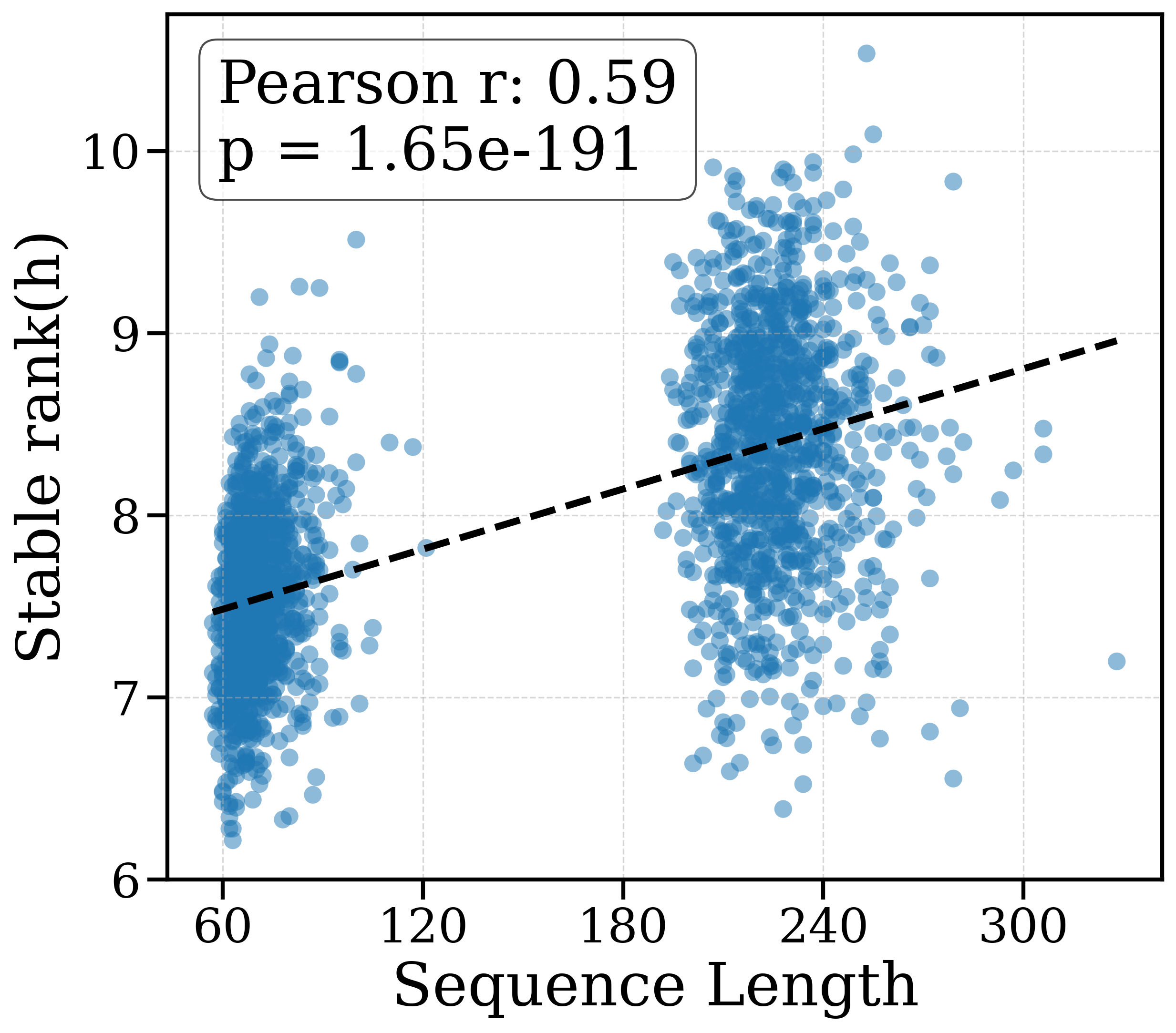}
        \caption{Hidden State Rank}
        \label{fig:sensitivity_hs}
    \end{subfigure}
    \begin{subfigure}[b]{0.3\textwidth}
        \centering
        \includegraphics[width=\textwidth]{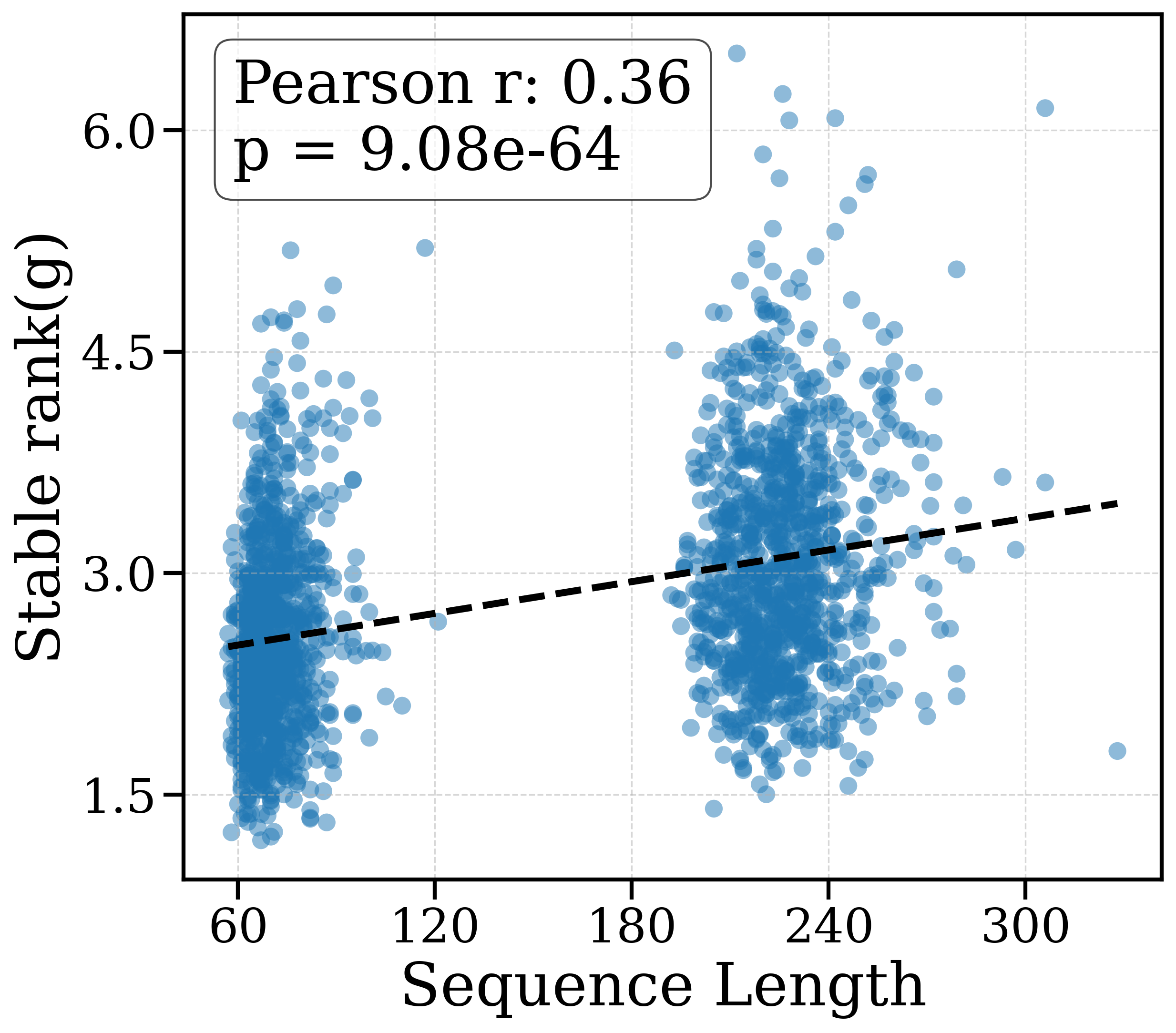}
        \caption{Stable Rank of Gradient}
        \label{fig:sensitivity_grad}
    \end{subfigure}
    \begin{subfigure}[b]{0.3\textwidth}
        \centering
        \includegraphics[width=\textwidth]{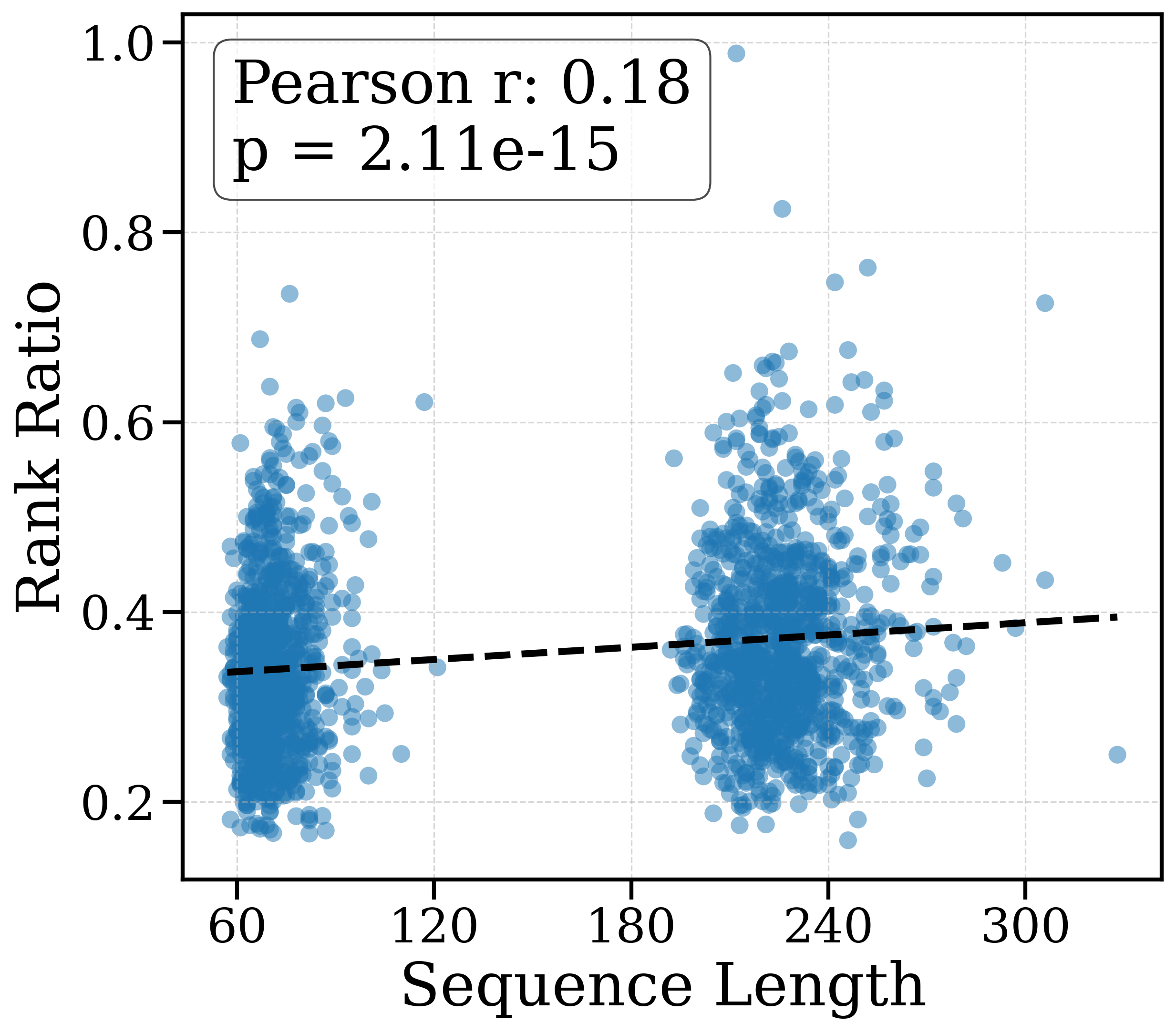}
        \caption{Rank Ratio}
        \label{fig:sensitivity_norm}
    \end{subfigure}
\end{center}
\caption{Pearson correlation between the input sequence length and the values of different rank-based metrics. The internals are extracted from Llama3.1-8B-Instruct model with HotpotQA as input. The rank ratio is the most robust metric to input length variation.} 
\label{fig:token_length_sensitivity}
\end{figure}

\textbf{Rank ratio as implicit normalization.} To quantify the knowledge gap, we introduce the Gradient Subspace Rank Ratio, which evaluates the proportion of effective required updates within the activated parameters. The gradient subspace rank ratio is computed as  ${srank(\mathcal{C}_g)}/{srank(\mathcal{C}_h)},$ 
where $srank(\mathcal{C}_h)$ is computed analogously to Eq. (\ref{eq:srank}).
More importantly, since longer inputs simultaneously expand both the gradient and hidden state subspaces, their ratio remains stable, acting as a natural normalization against input length variability.
To verify the justification, we calculate the correlation between metrics and the different input sequence lengths (results in Figure~\ref{fig:token_length_sensitivity}). Specifically, we select the queries from the HotpotQA~\citep{hqa} dataset that can't be correctly answered even when the model is fed with retrieved external documents (longer inputs shown in the x-axis). We observe that the rank of hidden states (a) is sensitive to the changes of the input length, with 0.59 Pearson correlation. The rank ratio (c) is the most robust (0.18), with a significant improvement over the stable rank (0.36) of the gradient (b).

\subsection{Supervised Probe based on Rank Ratio}
\label{subsec: probe}
To capture the dynamics of gradients across different layers, we gather an input feature consisting of $[\text{RankRatio}_1, \text{RankRatio}_2, \dots, \text{RankRatio}_L]$. This feature vector characterizes how ratio of required to activated knowledge evolves across layers, providing a richer propagation trajectory than scalar value-based thresholding. Empirical evidence in Table~\ref{tab:ablation_results} validates that layer-wise features are more indicative of knowledge sufficiency than such threshold-based alternatives.

To train the supervised probe, we construct a dataset of paired inputs consisting of layer-wise feature vectors and binary labels, where each label indicates whether the model can correctly answer the given query. In practice, for each query, we sample multiple responses and assign a positive label (answerable) if the accuracy across samples exceeds an upper threshold,
and a negative label (unanswerable) otherwise. Further details are in \S~\ref{app:probe}.




\section{Experiments}
\subsection{Experimental Settings}
\textbf{Models and dataset.} We use three backbone \textit{models}, i.e., Llama3.1-8b~\citep{llama3}, Qwen2.5-7b~\citep{qwen} and Gemma2-9b~\citep{gemma}. Evaluation are conducted across six datasets, including math (GSM8K~\citep{gsm8k}, MATH~\citep{math}), MMLU~\citep{mmlu}, and free-from QA: NQ~\citep{nq}, TQA~\citep{tqa} and HotpotQA~\citep{hqa}. 

\textbf{Baselines.} We compare with both verbalization, i.e., Judge \citep{black_priorijudge} and white-box, including token-probability and hidden states. P-Entropy \citep{white_predentropy} calculates the cumulative token-level entropy over the generated sequence. Internal Confidence (IC)~\citep{white_quncertainty} is unsupervised, mapping the last layer hidden state of the last input token to a confidence score $P(\text{YES})$, and Align-P~\citep{white_probe} is a supervised method by utilizing the intermediate hidden states before response generation.

\textbf{Metrics.} 
Following \citet{white_tokenconsistency} and \citet{white_quncertainty,black_semantic_uncertainty}, we utilize Accuracy (Acc) to calculate the proportion of samples where the predicted answerability aligns with the actual task performance; Area Under the Receiver Operating Characteristic (AUROC) \citep{auroc} as a robust evaluation insensitive to the class imbalance.~\footnote{Implementation details are shown in Appendix~\ref{app:inplement_details}.}
We focus on correctness-discrimination metrics rather than calibration metrics, as our objective is to identify knowledge boundaries instead of calibrating confidence. Further discussion and empirical comparisons with UQ calibration are provided in Appendix~\ref{sec:appendix_uq_vs_kbd}.

\subsection{Evaluation results of knowledge gap detection}

\textbf{Results for original queries.}
The knowledge gap detection for original queries in the evaluation benchmarks is shown in Table~\ref{tab:main_results}. Overall, our proposed $\modelname_\text{pre}$ and $\modelname_\text{pos}$ consistently outperform all baselines. Notably, $\modelname_\text{pre}$, which requires no response generation, achieves the best performance in 19 out of 36 settings, underscoring the strength of gradient-based signals as a lightweight yet effective indicator of knowledge sufficiency. Among the baselines, Align-P achieves the most competitive accuracy as a supervised probe, but suffers from severe AUROC degradation on complex datasets such as MMLU and HQA. 
Notably, the effectiveness of our two methods aligns with task complexity. For single-hop QA tasks, i.e., NQ and TQA, $\modelname_\text{pre}$ achieves the best with only 
the query-related information. For HQA and difficult reasoning datasets (MMLU, GSM8K and MATH), those requiring Chain-of-thought reasoning capabilities, $\modelname_\text{pos}$ can leverage its token-level information in the generated response to capture the procedure knowledge confidence. To further demonstrate the advantage of $\modelname$ over other recent representation-probing and reasoning-chain baselines (such as CoE~\citep{CoE} and SAPLMA~\citep{saplma_reference_key}), we provide an extended baseline comparison on our primary model backbone in Appendix~\ref{supp:extended_baselines}. The results show that $\modelname$ consistently achieves superior accuracy and AUROC.
\begin{table*}[h]
\centering
\resizebox{0.95\textwidth}{!}{
\begin{tabular}{p{0.5cm}l cccccccc cccc}
\toprule
\multirow{2}{*}{\textbf{}} & \multirow{2}{*}{\textbf{Method}} & \multicolumn{2}{c}{\textbf{NQ}} & \multicolumn{2}{c}{\textbf{TQA}} & \multicolumn{2}{c}{\textbf{HQA}} & \multicolumn{2}{c}{\textbf{MMLU}} & \multicolumn{2}{c}{\textbf{GSM8K}} & \multicolumn{2}{c}{\textbf{MATH}} \\
\cmidrule(lr){3-4} \cmidrule(lr){5-6} \cmidrule(lr){7-8} \cmidrule(lr){9-10} \cmidrule(lr){11-12} \cmidrule(lr){13-14}
& & Acc & AUROC & Acc & AUROC & Acc & AUROC & Acc & AUROC & Acc & AUROC & Acc & AUROC \\
\midrule
& Judge  &0.523 &- & 0.652&- & 0.536&- &0.519 &- &0.639 &- &0.501 &- \\
& P-Entropy &0.616 & 0.643&0.645 & 0.691& 0.596& 0.607&0.701 &0.728 & 0.643& 0.496& 0.661& 0.704\\
& IC & 0.640&0.669 & 0.661& 0.715& 0.747& 0.651& 0.504& 0.651& 0.518& 0.639& 0.575& 0.627\\
& Align-P & 0.771& 0.793& \underline{0.784}& \underline{0.855}& 0.737& 0.806 & \textbf{0.759} & 0.506 & 0.692& 0.766& 0.753 & 0.607 \\
\rowcolor{tablewhite} \multicolumn{1}{c}{\cellcolor{white}} & $\modelname_{\text{pos}}$ & \underline{0.794} & \underline{0.841} & 0.713 & 0.774 & \textbf{0.778} & \textbf{0.857} & \underline{0.727}& \textbf{0.778}& \textbf{0.836}& \textbf{0.893}&\textbf{0.891} &\textbf{0.962}\\
\rowcolor{tablewhite} \multicolumn{1}{c}{\cellcolor{white}\multirow{-6}{*}{\rotatebox{90}{\textbf{Llama3.1-8b}}}} & $\modelname_{\text{pre}}$ & \textbf{0.830} & \textbf{0.888} & \textbf{0.813} & \textbf{0.895} & \underline{0.754} & \underline{0.810} & 0.703 & \underline{0.751}& \underline{0.714}& \underline{0.767}&\underline{0.770} & \underline{0.841}\\
\midrule
& Judge &0.671 &- & 0.601&- & 0.536&- &0.490 &- &0.471 &- &0.662 &- \\
& P-Entropy & 0.596& 0.616 & 0.611& 0.649& 0.543&0.541 & 0.713 &0.678 &0.627 & 0.541& 0.702& 0.542\\
& IC & 0.625& 0.670& 0.607& 0.693& 0.612& 0.602& 0.502& 0.536& 0.603& 0.637& 0.676& 0.722\\
& Align-P &0.680 & 0.704 & 0.705 & \underline{0.833} & \underline{0.752} & 0.745 &\textbf{0.755} & 0.476 & 0.645 & 0.669 & 0.747& 0.797\\
\rowcolor{tablewhite} \multicolumn{1}{c}{\cellcolor{white}} & $\modelname_{\text{pos}}$ & 0.745& \underline{0.802}& 0.719& 0.776& \textbf{0.786}& \textbf{0.866}& 0.734 & \textbf{0.758}& \textbf{0.719}&\underline{0.772} & \textbf{0.879}&\textbf{0.921} \\
\rowcolor{tablewhite} \multicolumn{1}{c}{\cellcolor{white}\multirow{-6}{*}{\rotatebox{90}{\textbf{Qwen2.5-7b}}}} & $\modelname_{\text{pre}}$ & \textbf{0.781}& \textbf{0.814}& \textbf{0.795}& \textbf{0.862}& 0.715& \underline{0.778}& \underline{0.735}& \underline{0.688}& \underline{0.699}& \textbf{0.828}& \underline{0.804}&\underline{0.879} \\
\midrule
& Judge  & 0.574&- &0.692 &- &0.573 &- &0.472 &- & 0.513&- &0.400 &- \\
& P-Entropy &0.588 & 0.607& 0.631& 0.655&0.547 &0.538 & 0.702&0.765 & 0.739& \underline{0.794}& 0.691& 0.751 \\
& IC & 0.588& 0.650& 0.649& 0.698& 0.562& 0.655&0.505 &0.567 &0.507 &0.541 &0.715 &0.757 \\
& Align-P & 0.734& 0.730& 0.725& 0.799& 0.764& 0.855 & \textbf{0.782}& 0.534 &0.727 &0.776 & 0.613 & 0.636 \\
\rowcolor{tablewhite} \multicolumn{1}{c}{\cellcolor{white}} & $\modelname_{\text{pos}}$   & \underline{0.759}& \underline{0.823}& \underline{0.781}& \underline{0.818}& \underline{0.805}& \textbf{0.882}&\underline{0.768} & \textbf{0.838}& \textbf{0.776}& \textbf{0.821}& \textbf{0.871}&\textbf{0.934} \\
\rowcolor{tablewhite} \multicolumn{1}{c}{\cellcolor{white}\multirow{-6}{*}{\rotatebox{90}{\textbf{Gemma2-9b}}}} & $\modelname_{\text{pre}}$ & \textbf{0.814} & \textbf{0.907} & \textbf{0.832} & \textbf{0.932} & \textbf{0.815} & \underline{0.868} & 0.750 & \underline{0.809} & \underline{0.745} & 0.783 & \underline{0.794} & \underline{0.813} \\
\bottomrule
\end{tabular}
}
\caption{Main results on various datasets.}
\label{tab:main_results}
\end{table*}
\begin{figure}[htbp]
    \centering
    \includegraphics[width=0.99\linewidth,trim={0 0 0 6},clip]{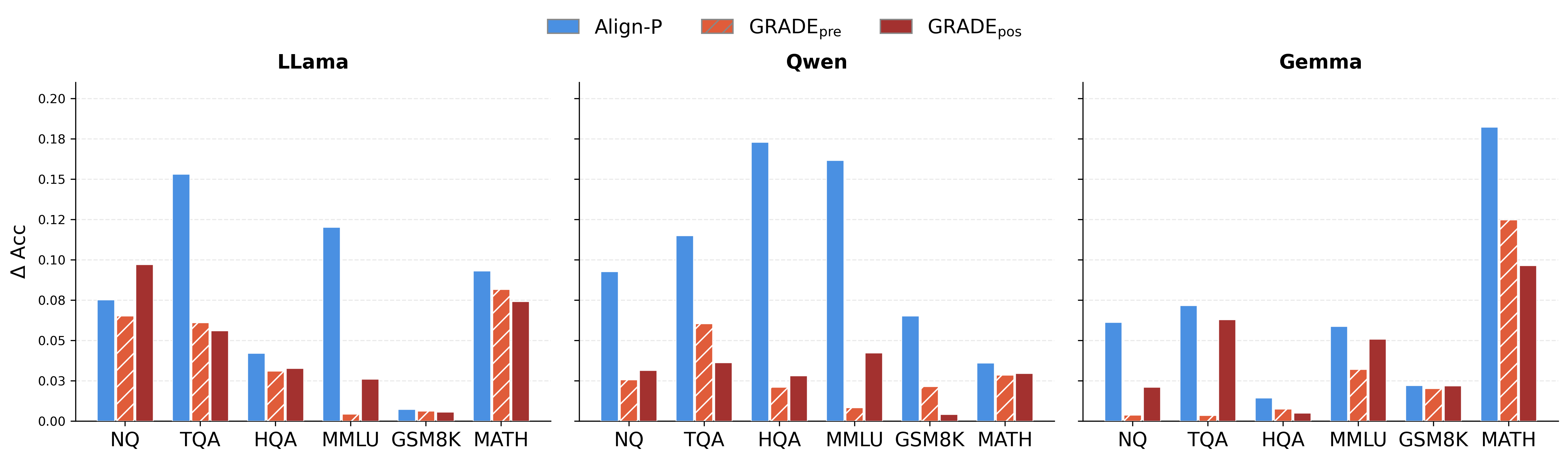}
    \caption{Relative change in detection accuracy ($\Delta Acc$) before and after input paraphrase. Smaller changes imply that the method is more robust to the perturbation.}
    \label{fig:perturbed_results}
\end{figure}
\paragraph{Robustness to input perturbation.} To evaluate the robustness of the methods against linguistic variation without confounding changes in task difficulty, we use Qwen3-30b-instruct to rephrase the original queries while preserving their core semantics. Crucially, to avoid selection bias and isolate pure linguistic robustness, we retain only query pairs where the LLM's answer correctness remains unchanged, thereby ensuring the ground-truth knowledge gap label is held fixed. A desirable knowledge gap detector should yield consistent predictions on these rephrased inputs. We quantify detector robustness by measuring the relative performance change, defined as $|\text{Acc}_{\text{rephrased}} - \text{Acc}_{\text{original}}|/{\text{Acc}_{\text{original}}}$.


The evaluation results on the perturbed inputs are presented in Figure~\ref{fig:perturbed_results}. The results clearly demonstrate that our proposed metrics, $\modelname_\text{pos}$ and $\modelname_\text{pre}$, are more robust to input perturbations compared to baselines. In contrast, Align-P exhibits notable performance fluctuations when input phrasing changes. This confirms that \modelname{} ignores superficial linguistic variations and reliably isolates intrinsic representational signals associated with actual knowledge gaps. Detailed relative change breakdowns for both Accuracy and AUROC, including IC (Internal Confidence), are provided in Appendix~\ref{supp:perturbation_auroc_ic}.

\begin{wraptable}{r}{0.6\textwidth}
\centering
\small
\resizebox{\linewidth}{!}{%
\begin{tabular}{ll ccc}
\toprule
\textbf{Direction} & \textbf{Method} & \textbf{NQ} & \textbf{TQA} & \textbf{HQA} \\
\midrule
\multicolumn{5}{l}{\textit{(a) Length Transfer: Long vs. Short}} \\
\multirow{2}{*}{Long $\rightarrow$ Short} 
 & Align-P & 0.687 / 0.733 & 0.747 / 0.826 & 0.654 / 0.726 \\
 & \textbf{\modelname{}} & \textbf{0.782} / \textbf{0.828} & \textbf{0.775} / \textbf{0.840} & \textbf{0.727} / \textbf{0.801} \\
\cmidrule(lr){1-5}
\multirow{2}{*}{Short $\rightarrow$ Long} 
 & Align-P & 0.731 / 0.806 & 0.769 / \textbf{0.829} & 0.684 / 0.790 \\
 & \textbf{\modelname{}} & \textbf{0.785} / \textbf{0.841} & \textbf{0.772} / 0.801 & \textbf{0.738} / \textbf{0.827} \\
\midrule
\multicolumn{5}{l}{\textit{(b) Context Transfer: With-Doc ($w$) vs. W/o-Doc ($w/o$)}} \\
\multirow{2}{*}{$w \rightarrow w/o$} 
 & Align-P & 0.716 / 0.799 & 0.682 / 0.780 & 0.751 / 0.821 \\
 & \textbf{\modelname{}} & \textbf{0.784} / \textbf{0.861} & \textbf{0.720} / \textbf{0.788} & \textbf{0.809} / \textbf{0.862} \\
\cmidrule(lr){1-5}
\multirow{2}{*}{$w/o \rightarrow w$} 
 & Align-P & 0.684 / 0.785 & 0.725 / \textbf{0.817} & 0.680 / 0.673 \\
 & \textbf{\modelname{}} & \textbf{0.750} / \textbf{0.843} & \textbf{0.752} / 0.809 & \textbf{0.762} / \textbf{0.842} \\
\bottomrule
\end{tabular}%
}
\caption{Cross-condition transfer performance (Acc / AUROC) across length and context variations.}
\label{tab:transfer_experiments}
\end{wraptable}
\paragraph{Robustness to length and context variations.} 
To isolate the impact of sequence length and external context, we conduct cross-condition transfer experiments by training and evaluating probes on symmetric subsets: (i) \textit{long vs. short} sequences, and (ii) \textit{with-document vs. without-document} settings. As presented in Table~\ref{tab:transfer_experiments}, \modelname{} consistently outperforms the baseline \textit{Align-P} across most cross-condition transfer directions on NQ, TQA, and HQA. Notably, when transferring from short to long sequences or across context settings (e.g., $w/o \rightarrow w$ on HQA, where \textit{Align-P}'s AUROC drops to $0.673$), \modelname{} maintains high classification accuracy and AUROC ($0.842$). These results confirm that \modelname{} effectively isolates intrinsic knowledge gaps and is robust to changes in sequence length and style.
\subsection{Cross-Dataset Transferability}
\begin{wrapfigure}{r}{0.7\textwidth}
    \centering
    \includegraphics[trim={12 12 10 10},clip, width=\linewidth]{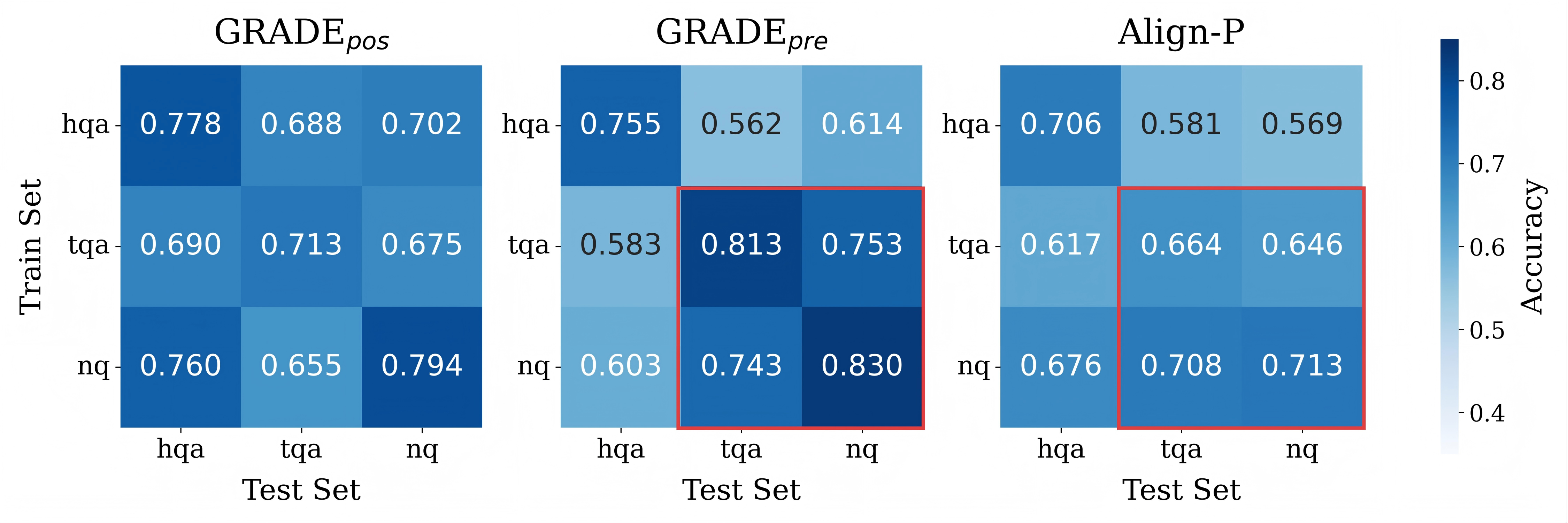}
    \caption{Cross-dataset generalization accuracy heatmaps. The results within the red box are transferred among similar complex questions (single-hop), i.e., between TQA and NQ.}
    \label{fig:ood_heatmaps}
\end{wrapfigure}

To assess the transferability, we evaluated \modelname{} by training on dataset A and testing on dataset B (see Figure~\ref{fig:ood_heatmaps}). By looking at the results at non-diagonal areas, we observe that $\modelname_\text{pos}$ demonstrates the most robust generalization, followed by $\modelname_\text{pre}$ and Align-P. Moreover, $\modelname_\text{pos}$ also demonstrates the best transfer patterns across different complexity datasets. Specifically, the transfer between similar complex datasets (TQA and NQ are single-hop) is shown within the red frame, and $\modelname{}_\text{pre}$ and Align-P show a significant degradation within this area. Noted that both $\modelname{}_\text{pre}$ and Align-P are methods using the generation logit before response sampling. In contrast, the superiority of $\modelname{}_\text{pos}$ suggests that sequence-level gradient dynamics along the generated response capture how required knowledge accumulates and interacts across tokens, making it more robust to distribution shifts between tasks of different complexities.

\section{Further Analysis}
We further analyze the effects of key designs of the proposed probes, such as modeling the gradient dynamics across different layers and the layer selections as probe inputs. We also show the generated fine-grained interpretation for the long-form response. 

\begin{figure}
    \centering
    \includegraphics[width=0.95\linewidth,trim={0 0 0 30},clip]{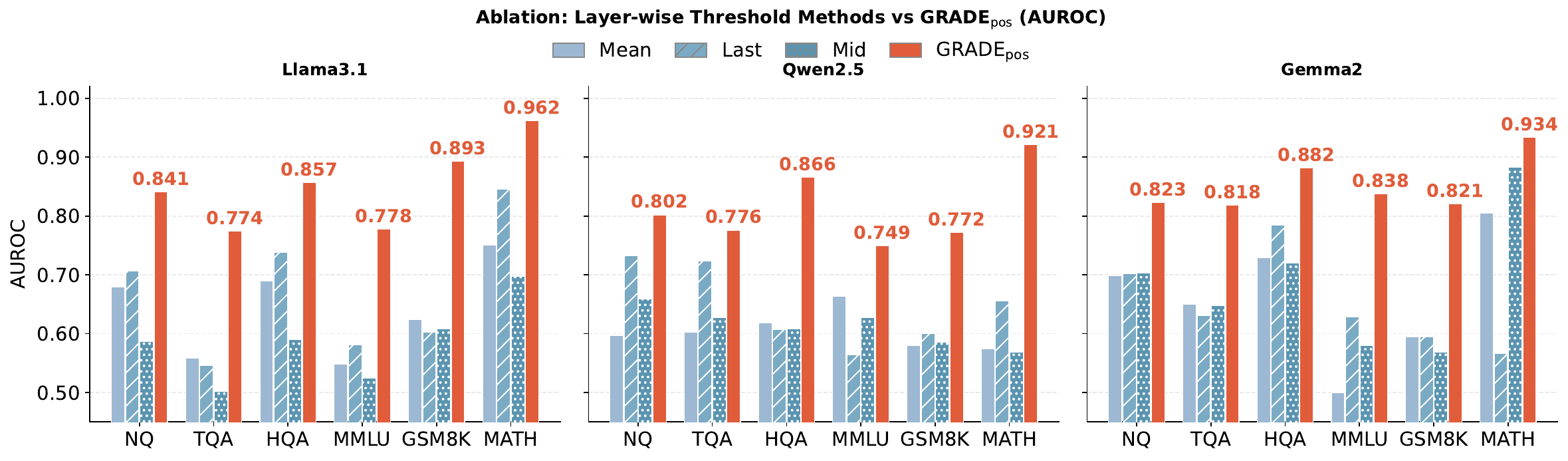}
    \caption{Comparison AUROC among detection thresholds (the mean, last-layer, and middle-layer rank ratio, respectively) versus \modelname{} (rank ratio patterns cross different layers).}
    \label{tab:ablation_results}
\end{figure}

\textbf{Effects of modeling the cross-layer dynamics.}
Some existing work suggests that layer-wise signals~\citep{white_probe,final_layer} are reliable indicators of model behaviour, motivating a simple thresholding approach: take the rank ratio from a specific layer, determine an answerable/unanswerable threshold on the training set, and apply it to the test set. We compare several such variants in Figure~\ref{tab:ablation_results}, where Mean, Last, and Mid denote using the average, last-layer, and middle-layer rank ratio as the thresholding signal, respectively. In contrast, $\modelname{}_\text{pos}$ takes the full layer-wise rank~\footnote{We ablate the layer selections and the results in Table~\ref{fig:ablation_layer} showing that taking all layer-wise rank ratio is the best-performing setup.} ratio vector as probe input, enabling it to learn cross-layer propagation patterns rather than relying on any single layer's signal. The results confirm the effectiveness of this design, with a clear and consistent performance gap over all thresholding baselines (Acc results are shown in Figure~\ref{fig:com_thres_acc}). We also show the rank ratios across different layers for both answerable and unanswerable queries in Figure~\ref{fig:layer_trend}: it shows that their absolute values for the two categories are not clearly distinguishable, but the dynamics across different layers differ.

\begin{wrapfigure}{r}{0.6\textwidth}
    \centering
    \includegraphics[width=\linewidth, trim={0 0 0 10},clip]{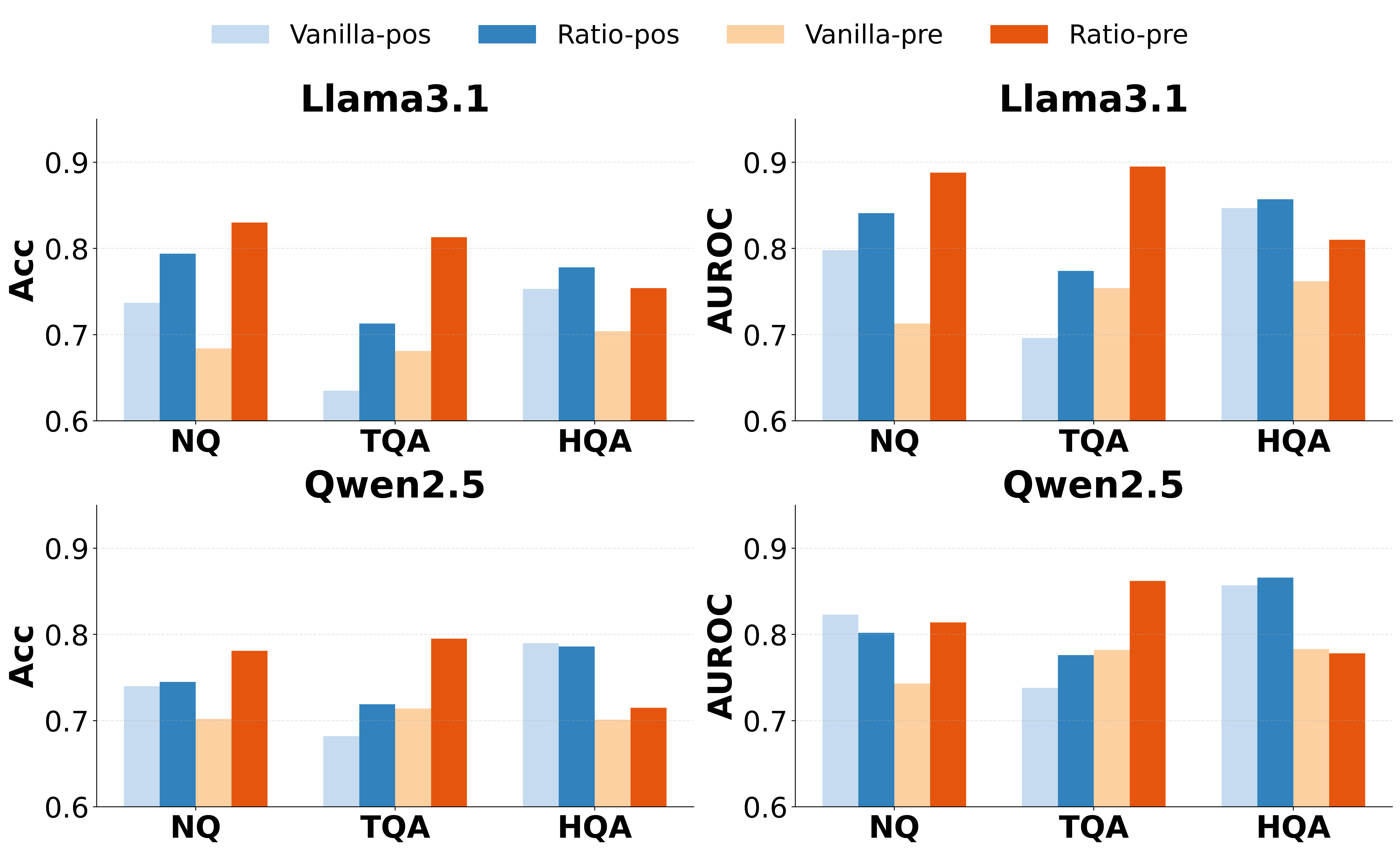}
    \caption{Ablation on the effectiveness of the rank ratio. We compare the proposed rank ratio with the vanilla gradient rank on Llama3.1 and Qwen2.5 across NQ, TQA, and HQA.}
    \label{tab:rank_ratio_ablation}
\end{wrapfigure}

\textbf{Ablation on Rank-Ratio vs. Vanilla Gradient Rank.}
To isolate the contribution of the rank-ratio normalization, we evaluate a baseline that directly uses the vanilla gradient rank vector as the probing feature without dividing by hidden-state ranks. As shown in Table~\ref{tab:rank_ratio_ablation}, the proposed rank-ratio vector outperforms raw gradient rank across most settings on both LLaMA and Qwen backbones. This confirms that normalizing the gradient dynamics against hidden-state capacity effectively reduces sequence-length interference. Nevertheless, the vanilla gradient rank itself achieves strong accuracy and AUROC, demonstrating that parameter-level gradient dynamics inherently provides a powerful discriminative signal for the detection of knowledge boundaries.

\paragraph{Interpretation.}
\label{subsec:results_explain}
As $\modelname{}_\text{pos}$ considers the information across all the generated tokens, we can thus leverage it to generate token-level interpretation about the knowledge sufficiency. Specifically, we use the $t$-th row-sum of $\mathcal{C}_g$ to identify the knowledge gap of token $t$. A high score indicates an epistemic shock, signifying a knowledge gap where the model requires large updates; a low score suggests that the token resides within the model's knowledge boundary, requiring minimal adjustment. To ensure a fair and consistent comparison, the scores are globally normalized across all samples. 

\begin{figure}[h]
    \centering
    \begin{subfigure}[b]{0.95\linewidth}
        \centering
        \includegraphics[width=\linewidth,trim={10 35 10 10},clip]{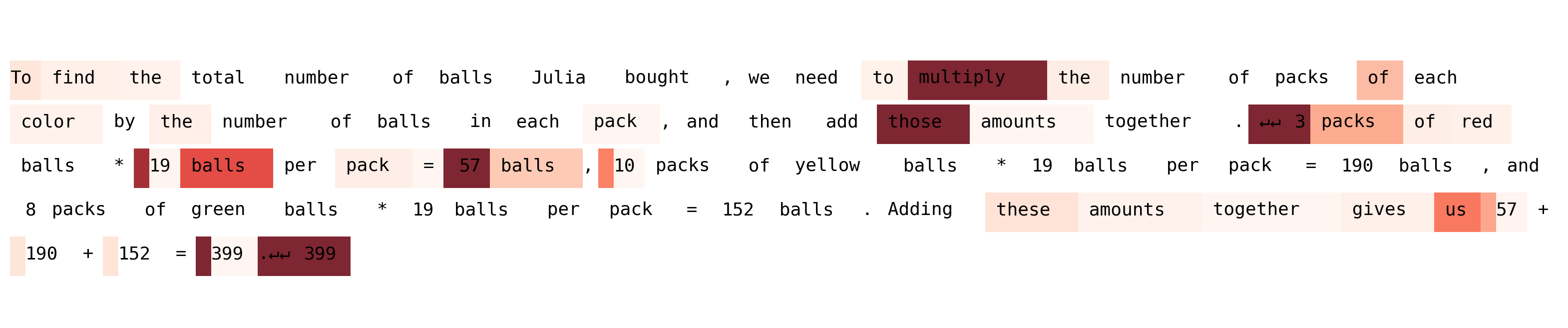}
        \caption{Correct reasoning outcome.}
        \label{fig:answer}
    \end{subfigure}
    \begin{subfigure}[b]{0.95\linewidth}
        \centering
        \includegraphics[width=\linewidth,trim={10 35 10 10},clip]{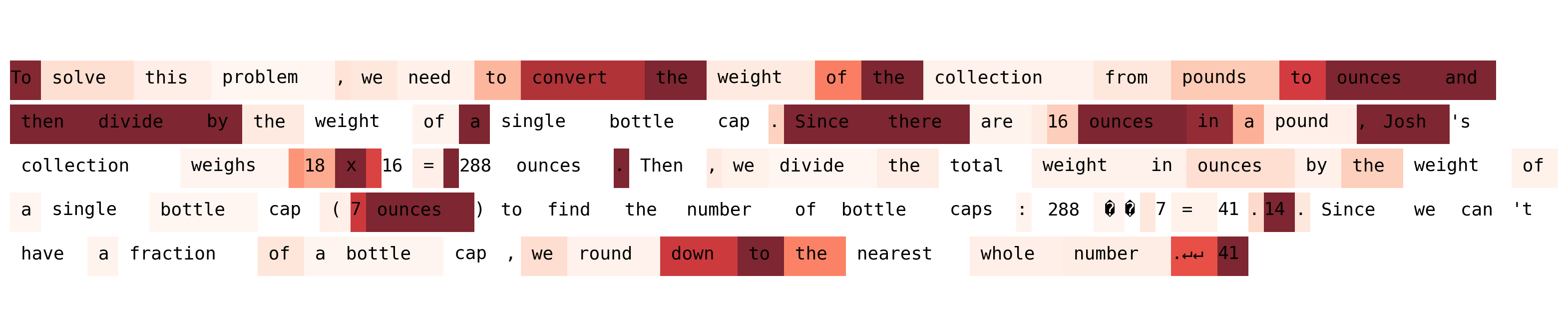}
        \caption{Incorrect reasoning sample.}
        \label{fig:not_answer_0}
    \end{subfigure}
    \caption{Token-level metric across correctly and incorrectly answered examples.}
    \label{fig:token_viz}
\end{figure}

We show the generated uncertainty maps for both correctly and incorrectly answered queries from the GSM8k dataset in Figure \ref{fig:answer}. We observe that the high scores are predominantly concentrated on critical logical anchors and mathematical operators, such as \textit{`multiply'} and the subsequent numeric result. 
This suggests that the model experiences momentary cognitive pressure only at pivotal decision points or computational junctions. Once these key steps are resolved, the scores rapidly decay, indicating that the rest of the reasoning chain resides safely within the internal knowledge. For the incorrect reasoning, we observe a rather diffuse pattern across the whole inputs, such as a high score is observed during the unit conversion process (e.g., \textit{`convert'}, \textit{`ounces'}), which persists throughout the following tokens. This pattern reflects a global epistemic collapse. 


\vspace{-2mm}
\section{Conclusion}
\vspace{-2mm}
We propose \modelname{} (\textbf{G}\textbf{R}\textbf{A}dient \textbf{D}ynamics for knowl\textbf{E}dge gap detection), a gradient-based approach that addresses the fundamental challenge of detecting knowledge gaps for a given query. Unlike hidden-state methods that capture activated knowledge richness, \modelname{} captures the query-specific knowledge structure reflected in parameter updates. Specifically, \modelname{} quantifies the rank ratio between the gradient and hidden state subspaces, and uses the cross-layer ratio tendency as input to train a gap detector. Validation across six benchmarks demonstrates superior robustness and accuracy over verbalised, probabilistic, and hidden-state baselines, and a case study further illustrates how the gradient chain yields interpretable explanations of knowledge gaps in long-form answers.

\section*{Acknowledgment}
This work was supported by the Start-up funding from King's College London, and the Engineering and Physical Sciences Research Council (No. EP/Y028805/1).

\bibliography{colm2026_conference}
\bibliographystyle{colm2026_conference}

\clearpage
\appendix
\section{Appendix}
\subsection{Implementation details}
\label{app:inplement_details}
\subsubsection{Data Setup}
For open-domain QA datasets (NQ, TriviaQA, HotpotQA), we evaluate both no-context and with-doc settings.
In the with-doc setting, we employ the Contriever \citep{contriever} dense retriever to retrieve relevant evidence passages. For each query, the top-1 retrieved document is selected and prepended to the question within the prompt. This document is inserted as an additional context block before the question to provide external evidence for answer generation.
\subsubsection{Knowledge Gap Labeling}
To determine answerability, we categorize queries based on their empirical accuracy across 10 samples for QA and multiple-choice tasks, and 5 samples for CoT reasoning. For the GSM8K dataset, we define the knowledge boundary using thresholds of 0.6 (answerable) and 0.4 (unanswerable), while other datasets utilize 0.8 and 0.2, respectively. As detailed in Table~\ref{tab:retained_ratio}, the vast majority of samples (76.9\%--92.4\%) are preserved across all datasets, confirming that the filtering process maintains dataset integrity without discarding excessive data. Evaluation is conducted using Exact Match (EM) for QA and multiple-choice tasks, whereas CoT tasks are evaluated based on the correctness of the final result extracted from the reasoning chain.
\begin{table}[htbp]
\centering
\begin{tabular}{lcccccc}
\toprule
\textbf{Dataset} & \textbf{NQ} & \textbf{TQA} & \textbf{HQA} & \textbf{MMLU} & \textbf{GSM8K} & \textbf{MATH} \\
\midrule
\textbf{Retained Ratio} & 91.5\% & 87.2\% & 92.4\% & 76.9\% & 87.8\% & 82.0\% \\
\bottomrule
\end{tabular}
\caption{Retained sample ratio across different benchmark datasets after empirical accuracy filtering ($\text{Acc} \ge 0.8$ for answerable and $\text{Acc} \le 0.2$ for unanswerable).}
\label{tab:retained_ratio}
\end{table}

\subsubsection{\modelname{}$_\text{pos}$ for Long-form Generation}
To evaluate $\modelname{}_{pos}$ in extended reasoning scenarios, we apply our gradient-based analysis to segmented sequences. Following the generation of a complete reasoning chain, we segment the response into discrete steps $\{\mathcal{S}_1, \mathcal{S}_2, \dots, \mathcal{S}_K\}$ based on punctuation markers (e.g., "."). For a given reasoning step $\mathcal{S}_k$, let $n_k = n_{\text{q}} + \sum_{j=1}^k |\mathcal{S}_j|$ be the cumulative length of the sequence up to the end of this step, where $n_{\text{q}}$ denotes the length of the input prompt. The loss function of the current step is: 
\begin{equation}
    \mathcal{L}_{pos}^{(k)} = - \sum_{t \in \mathcal{S}_k} \log p(y_t \mid \bm{x}_{<t}),
\end{equation}
where only the tokens within the current step $\mathcal{S}_k$ contribute. Let $\bm{h}^{(k)} \in \mathbb{R}^{n_k \times d_{\text{ff}}}$ denote the intermediate hidden states,with $\bm{g}^{(k)}$ representing the corresponding step-specific gradient. Finally, to derive the metric $\modelname{}_c$ for the complete sequence, we aggregate the step-specific stable rank ratios by averaging them across all $K$ steps:
\begin{equation}
    \modelname{}_{pos} = \frac{1}{K} \sum_{k=1}^{K} \frac{srank(\bm{g}^{(k)})}{srank(\bm{h}^{(k)})}.
\end{equation}

While $r_{G}$ captures the continuous dimensionality of the knowledge flux, it suffers from sensitivity to sequence length, since longer contexts inherently possess higher rank ceilings and introduce more baseline degrees of freedom. Let $\sigma_1 \ge \sigma_2 \ge \dots \ge \sigma_s > 0$ be the sorted eigenvalues of the hidden Gram matrix $H H^\top$. We define the effective rank of the hidden-state $r_{H}$ similarly:$$r_{H} = \sum_{i=1}^{s} \frac{\sigma_i}{\sigma_1}$$Finally, we propose the Gradient Subspace Rank Ratio (GSRR) to quantify the relative geometric alignment between the parameter update and the representation subspace:$$\mathrm{GSRR} = \frac{r_{G}}{r_{H}}.$$

\subsubsection{Probe Architecture}
\label{app:probe}
The alignment probe is implemented as a 5-layer feed-forward neural network designed to progressively compress input features into a scalar probability ($num\_layers \to 256 \to 128 \to 64 \to 32 \to 1$). Each hidden block consists of a linear transformation followed by Batch Normalization, a Leaky ReLU activation (negative slope $= 0.01$), and a Dropout layer ($p = 0.5$) to enhance generalization and prevent overfitting.
We employ Kaiming uniform initialization for all linear weights and initialize batch normalization parameters to unit scale and zero bias. The final layer utilizes a Sigmoid activation to map the output to $[0, 1]$.

\subsubsection{Probe Training}
\label{app:probe_train}
We implement the training pipeline for the diagnostic probe using PyTorch. To ensure strict reproducibility across all experiments, the global random seed is fixed to 42. The probe is trained for a maximum of 100 epochs with a batch size of 64. For optimization, we initialize the learning rate at $5 \times 10^{-3}$ and apply a weight decay of $1 \times 10^{-5}$ to regularize the network and mitigate overfitting. To facilitate stable convergence and avoid local minima, we incorporate a learning rate scheduler that dynamically reduces the learning rate by a factor of 0.75 if no improvement is observed over a patience window of 5 epochs. Finally, during the evaluation phase, we adopt a standard decision threshold of 0.5 to binarize the probe's continuous output into a definitive prediction.

\subsubsection{Paraphrase Generation}
\label{app_para}
For each query in datasets, we generate a semantically equivalent paraphrase using Qwen3-30B-A3B-Instruct-2507 \citep{qwen3}. To ensure the perturbations only affect the linguistic form without altering the core factual or logical requirements, we employ a zero-shot prompting strategy. The system is provided with the following instruction:

System Prompt: "You are a helpful assistant. Please paraphrase the following question accurately without changing its original meaning or entity names. Output ONLY the paraphrased question."

\subsubsection{Calculation of AUROC}
The Area Under the Receiver Operating Characteristic curve (AUROC) is a threshold-independent metric utilized to evaluate the discriminative capability of our gradient-based probing classifier. It is particularly advantageous in our setting as it remains robust against class imbalance—a common scenario when evaluating the proportion of answerable versus unanswerable queries in LLMs.

Let $\mathcal{D}^+$ represent the set of positive samples (i.e., queries the model answers correctly) and $\mathcal{D}^-$ represent the set of negative samples (i.e., queries exposing a knowledge gap). Let $s(x) \in [0, 1]$ denote the predicted confidence score output by our probe for a given query $x$. For any decision threshold $\tau \in [0, 1]$, the True Positive Rate (TPR) and False Positive Rate (FPR) are formally defined as:$$\text{TPR}(\tau) = \frac{|\{x \in \mathcal{D}^+ \mid s(x) \ge \tau\}|}{|\mathcal{D}^+|}$$$$\text{FPR}(\tau) = \frac{|\{x \in \mathcal{D}^- \mid s(x) \ge \tau\}|}{|\mathcal{D}^-|}.$$The ROC curve is constructed by plotting $\text{TPR}(\tau)$ against $\text{FPR}(\tau)$ as the threshold $\tau$ varies from 1 down to 0. The AUROC is geometrically defined as the integral of this curve:$$\text{AUROC} = \int_{0}^{1} \text{TPR}(\tau) \, d(\text{FPR}(\tau)).$$

\subsection{Dataset Descriptions}
GSM8K \citep{gsm8k} is a collection of 8,500 high-quality grade-school math word problems designed to evaluate models’ abilities in multi-step arithmetic reasoning. The dataset focuses on interpretable numerical reasoning rather than symbolic manipulation, making it a standard benchmark for assessing mathematical problem-solving in natural language.

MATH \citep{math} contains more than 12,000 challenging competition-level mathematics problems covering algebra, geometry, number theory, combinatorics, and calculus. Each problem includes step-by-step solutions, enabling rigorous evaluation of advanced mathematical reasoning and multi-step derivation abilities.

The Massive Multitask Language Understanding (MMLU) benchmark \citep{mmlu} consists of 57 multiple-choice subjects spanning STEM, humanities, social sciences, and professional domains. It assesses models’ world knowledge and problem-solving skills across a broad range of disciplines, making it one of the most comprehensive evaluations of general knowledge reasoning.

Natural Questions (NQ) \citep{nq} comprises real anonymized queries issued to the Google search engine, paired with evidence passages from Wikipedia. It evaluates open-domain question answering, requiring systems to locate and extract relevant information from long documents under realistic information-seeking scenarios.

TriviaQA (TQA) \citep{tqa} is a large-scale open-domain question answering dataset consisting of over 95,000 question–answer pairs authored by trivia enthusiasts. Each question is paired with evidence documents from the web or Wikipedia, enabling evaluation of retrieval-based QA, reasoning over long contexts, and robust answer extraction.

HotpotQA (HQA) \citep{hqa} is an open-domain QA dataset featuring multi-hop reasoning questions that require retrieving and integrating information from multiple Wikipedia articles. It supports both extractive and abstractive answers and provides supporting facts, enabling evaluation of interpretable multi-step reasoning.

\subsection{Baselines Implementation}
\label{app:baseline}
\textbf{Judge:} The Prior Judge baseline evaluates epistemic uncertainty by directly prompting the Large Language Model (LLM) to explicitly state whether it possesses the knowledge to answer a question. 

(1) In the closed-book setting, we use the prompt: "Are you sure to accurately answer the following question based on your internal knowledge? If yes, you should give a short answer with one or a few words; if no, you should answer 'Unknown'. Question: {question} Answer: ". 

(2) For the Retrieval-Augmented Generation (RAG) setting, the prompt is adapted to: "Given the following information: {context} Can you answer the following question based on the given information or your internal knowledge, if yes, you should give a short answer with one or few words, if no, you should answer 'Unknown'. Question: {question} Answer:". 

During inference, we enforce greedy decoding (temperature 0.0) and restrict the maximum generation length to 16 tokens to ensure deterministic and concise self-assessment. The generated text is then mapped to a binary confidence indicator $\hat{c}\in\{0,1\}$. If the response contains the keyword "Unknown" (case-insensitive), we assign a confidence score of $\hat{c}=0$, indicating a lack of knowledge. Conversely, any other generated short answer yields $\hat{c}=1$. 

\textbf{P-entropy:} The Predictive Entropy baseline quantifies epistemic uncertainty by aggregating the token-level Shannon entropy across the entire generated response. For a given instruction of length $L_{instr}$ and a generated response of length $L_{resp}$, we extract the probability distribution $P_M(x_t | x_{<t})$ over the model's full vocabulary $V$ at each generation step $t$. The total predictive entropy $H$ is then calculated as the sum of the step-wise entropies:$$H = - \sum_{t=L_{instr}+1}^{L_{instr}+L_{resp}} \sum_{x_t \in V} P_M(x_t | x_{<t}) \log P_M(x_t | x_{<t}).$$

\textbf{IC (Internal Confidence): }We follow the training-free implementation of Internal Confidence. Rather than performing any data-driven search for an optimal center, we use a fixed decision center positioned at the last layer and the last token position. The confidence score is computed by aggregating the log-probability of $P(\text{YES})$ from this fixed center alongside its immediate spatial (neighboring tokens) and depth (neighboring layers) neighborhoods.

For a prompt of length $N$ and a model with $L$ layers, let the hidden state at layer $l$ and token position $n$ be $h_n^{(l)}$. Instead of relying solely on the final output token, these intermediate representations are extracted. Each hidden state is individually passed through the model's language modeling head to isolate the probability of predicting the "Yes" target token, denoted as $P(\text{YES} | h_n^{(l)})$. To compute the final continuous confidence score, these intermediate probabilities are aggregated using a weighted average. This aggregation is built around an optimal "decision center"—the specific layer and token position where the model most effectively separates answerable from non-answerable queries. The Internal Confidence ($IC$) is formally defined as:$$IC(h) = \sum_{n=1}^N \sum_{l=1}^L w_n^{(l)} P(\text{YES} | h_n^{(l)})$$where $w_n^{(l)}$ denotes the weight assigned to the hidden representation $h_n^{(l)}$. These weights are determined by a positional attention mechanism that applies a distance-based decay originating from the decision center, placing the highest emphasis on the final $k$ tokens and the deepest layers of the model.

\subsection{Additional experiment results}
\subsubsection{Uncertainty Quantification vs. Knowledge Boundary Detection}
\label{sec:appendix_uq_vs_kbd}

While both Uncertainty Quantification (UQ) and Knowledge Boundary Detection (KBD) utilize binary label correctness for evaluation, they differ in core assumptions and target objectives:

\begin{itemize}[leftmargin=*]
    \item \textbf{Underlying Assumptions and Failure Modes:} Standard UQ methods assume that a model's internal confidence (or entropy/consistency) aligns with prediction correctness. However, LLMs frequently suffer from severe overconfidence—generating consistent yet incorrect answers for queries outside their knowledge base, leading traditional UQ metrics to misinterpret certainty for correctness.
    \item \textbf{Evaluation Focus (Calibration vs. Discrimination):} Standard UQ focuses on \textit{calibration} (e.g., Expected Calibration Error), measuring how well confidence scores align with accuracy. In contrast, Knowledge Boundary Detection focuses explicitly on \textit{correctness discrimination}—identifying whether required latent knowledge is present in the model, regardless of its output confidence.
\end{itemize}

\paragraph{Empirical Evaluation on "Confidently Wrong" Cases.} To compare these approaches when confidence signals fail, we construct a hard evaluation set of \textbf{500 "confidently wrong" samples} where the LLM repeatedly generates identical incorrect answers across multiple independent sampling passes. A robust knowledge boundary detector should identify these queries as \textit{unanswerable} despite high confidence. 

As shown in Table~\ref{tab:confidently_wrong_results}, unsupervised UQ baselines perform poorly: \textit{Self-consistency} \citep{sc_reference_key} achieves $0.0\%$ accuracy due to identical generation paths, and \textit{P-entropy} \citep{white_predentropy} achieves only $32.8\%$. Supervised \textit{Align-P} \citep{white_probe} reaches $67.5\%$. In contrast, \modelname{} achieves the highest accuracy of \textbf{72.6\%}, demonstrating that rank ratio analysis on projected gradient dynamics effectively detects knowledge gaps under overconfidence.

\begin{table}[h]
\centering
\small
\begin{tabular}{llc}
\toprule
\textbf{Category} & \textbf{Method} & \textbf{Accuracy (\%)} \\
\midrule
\multirow{2}{*}{Unsupervised UQ} 
 & P-entropy \citep{white_predentropy} & 32.8 \\
 & Self-consistency \citep{sc_reference_key} & 0.0 \\
\midrule
Supervised Baseline 
 & Align-P \citep{white_probe} & 67.5 \\
\midrule
\rowcolor{tablewhite}
\textbf{Ours} & \textbf{\modelname{}} & \textbf{72.6} \\
\bottomrule
\end{tabular}
\caption{Detection accuracy on 500 "confidently wrong" queries where the LLM produces consistently incorrect answers. Higher accuracy indicates superior capability in identifying knowledge gaps despite high output confidence.}
\label{tab:confidently_wrong_results}
\end{table}
\subsubsection{Extended Baseline Evaluations}
\label{supp:extended_baselines}
To further evaluate $\text{GRADE}_{\text{pre}}$ against a broader range of internal-representation probing and self-evaluation approaches, we expand our empirical comparisons by including three additional baselines on LLaMA-3-8B across all six benchmark datasets:
(1) \textbf{CoE}~\citep{CoE}, which models progressive layer-wise embeddings as a latent reasoning trajectory;
(2) \textbf{SAPLMA}~\citep{saplma_reference_key}, a supervised probing method targeting final-token hidden activations from generated responses; and 
(3) \textbf{Self-Consistency}~\citep{sc_reference_key}, a widely used non-probing self-evaluation baseline measuring generation stability across multiple sample paths.

As shown in Table~\ref{tab:updated_baselines}, $\text{GRADE}_{\text{pre}}$ consistently achieves superior accuracy and AUROC compared to these baselines on almost all tasks. While CoE and SAPLMA show competitive accuracy on individual benchmarks (e.g., MMLU), their AUROC performance is significantly lower than $\text{GRADE}_{\text{pre}}$, indicating weaker overall ranking capability for knowledge boundary detection. These results confirm that gradient-subspace rank features offer a more reliable discriminative signal than standard hidden-state probing methods or output repetition counts.
\begin{table*}[htbp]
\centering
\resizebox{0.95\textwidth}{!}{
\begin{tabular}{l cccccccc cccc}
\toprule
\multirow{2}{*}{\textbf{Method}} & \multicolumn{2}{c}{\textbf{NQ}} & \multicolumn{2}{c}{\textbf{TQA}} & \multicolumn{2}{c}{\textbf{HQA}} & \multicolumn{2}{c}{\textbf{MMLU}} & \multicolumn{2}{c}{\textbf{GSM8K}} & \multicolumn{2}{c}{\textbf{MATH}} \\
\cmidrule(lr){2-3} \cmidrule(lr){4-5} \cmidrule(lr){6-7} \cmidrule(lr){8-9} \cmidrule(lr){10-11} \cmidrule(lr){12-13}
& Acc & AUROC & Acc & AUROC & Acc & AUROC & Acc & AUROC & Acc & AUROC & Acc & AUROC \\
\midrule
CoE & 0.790 & 0.626 & 0.605 & 0.634 & 0.643 & 0.511 & \textbf{0.724} & 0.554 & 0.617 & 0.666 & 0.693 & 0.537 \\
SAPLMA & 0.716 & 0.788 & 0.712 & 0.784 & 0.729 & 0.796 & 0.664 & 0.523 & 0.619 & 0.602 & 0.716 & 0.790 \\
self-consistency & 0.572 & 0.830 & 0.698 & 0.791 & 0.535 & \textbf{0.825} & 0.583 & 0.602 & 0.372 & 0.423 & 0.491 & 0.500 \\
\midrule
\rowcolor{tablewhite} $\modelname_{pre}$ & \textbf{0.830} & \textbf{0.888} & \textbf{0.813} & \textbf{0.895} & \textbf{0.754} & 0.810 & 0.703 & \textbf{0.751} & \textbf{0.714} & \textbf{0.767} & \textbf{0.770} & \textbf{0.841} \\
\bottomrule
\end{tabular}
}
\caption{Performance comparison with updated baselines.}
\label{tab:updated_baselines}
\end{table*}

\subsubsection{Detailed Perturbation Analysis Including AUROC and IC Baseline}
\label{supp:perturbation_auroc_ic}

To provide a comprehensive evaluation of detector robustness under input perturbation, we extend our comparison to include the IC baseline as well as the relative change in AUROC performance. Table~\ref{tab:perturbation_full_results} presents the relative performance fluctuations ($\%$) for both Accuracy and AUROC before and after query paraphrasing.

\begin{table}[h]
\centering
\scriptsize
\resizebox{0.8\linewidth}{!}{%
\begin{tabular}{l cccccc}
\toprule
\textbf{Method} & \textbf{NQ} & \textbf{TQA} & \textbf{HQA} & \textbf{MMLU} & \textbf{GSM8K} & \textbf{MATH} \\
\midrule
\multicolumn{7}{l}{\textit{(a) Accuracy Relative Change (\%)}} \\
IC & 21.7 & \textbf{3.9} & 3.8 & 3.3 & 16.2 & 21.7 \\
Align-P & 7.5 & 15.8 & 4.2 & 11.9 & 0.7 & 9.3 \\
\rowcolor{tablewhite} $\text{\modelname{}}_{\text{pos}}$ & 9.6 & 5.2 & 3.2 & 2.6 & \textbf{0.5} & 7.4 \\
\rowcolor{tablewhite} $\text{\modelname{}}_{\text{pre}}$ & \textbf{6.5} & 6.1 & \textbf{3.1} & \textbf{0.4} & 0.6 & \textbf{8.1} \\
\midrule
\multicolumn{7}{l}{\textit{(b) AUROC Relative Change (\%)}} \\
IC & 4.0 & \textbf{2.1} & \textbf{0.4} & 2.0 & 1.8 & 20.5 \\
Align-P & 3.6 & 17.3 & 5.9 & 27.6 & 2.3 & 42.0 \\
\rowcolor{tablewhite} $\text{\modelname{}}_{\text{pos}}$ & 3.7 & 4.7 & 3.3 & \textbf{1.7} & \textbf{1.2} & \textbf{3.5} \\
\rowcolor{tablewhite} $\text{\modelname{}}_{\text{pre}}$ & \textbf{3.0} & 10.1 & 8.3 & 1.8 & 1.8 & 6.3 \\
\bottomrule
\end{tabular}%
}
\caption{Relative performance fluctuations ($\%$) under input perturbation for Accuracy (a) and AUROC (b). Smaller values indicate higher robustness to linguistic variations.}
\label{tab:perturbation_full_results}
\end{table}

\paragraph{Analysis.} Across the vast majority of benchmarks, $\text{\modelname{}}$ models exhibit consistently smaller performance fluctuations in both Accuracy and AUROC compared to representation- and logit-based baselines like IC and Align-P. Specifically, while Align-P suffers severe AUROC degradation on complex reasoning datasets (e.g., $27.6\%$ drop on MMLU and $42.0\%$ on MATH), $\text{\modelname{}}$ maintains exceptional stability (less than $2\%$ shift on MMLU), further proving its resilience against superficial syntactic changes.

\subsubsection{Rank Calculation is sensitive to threshold selection.}
\begin{figure}[t]
  \centering
  \includegraphics[width=0.5\linewidth]{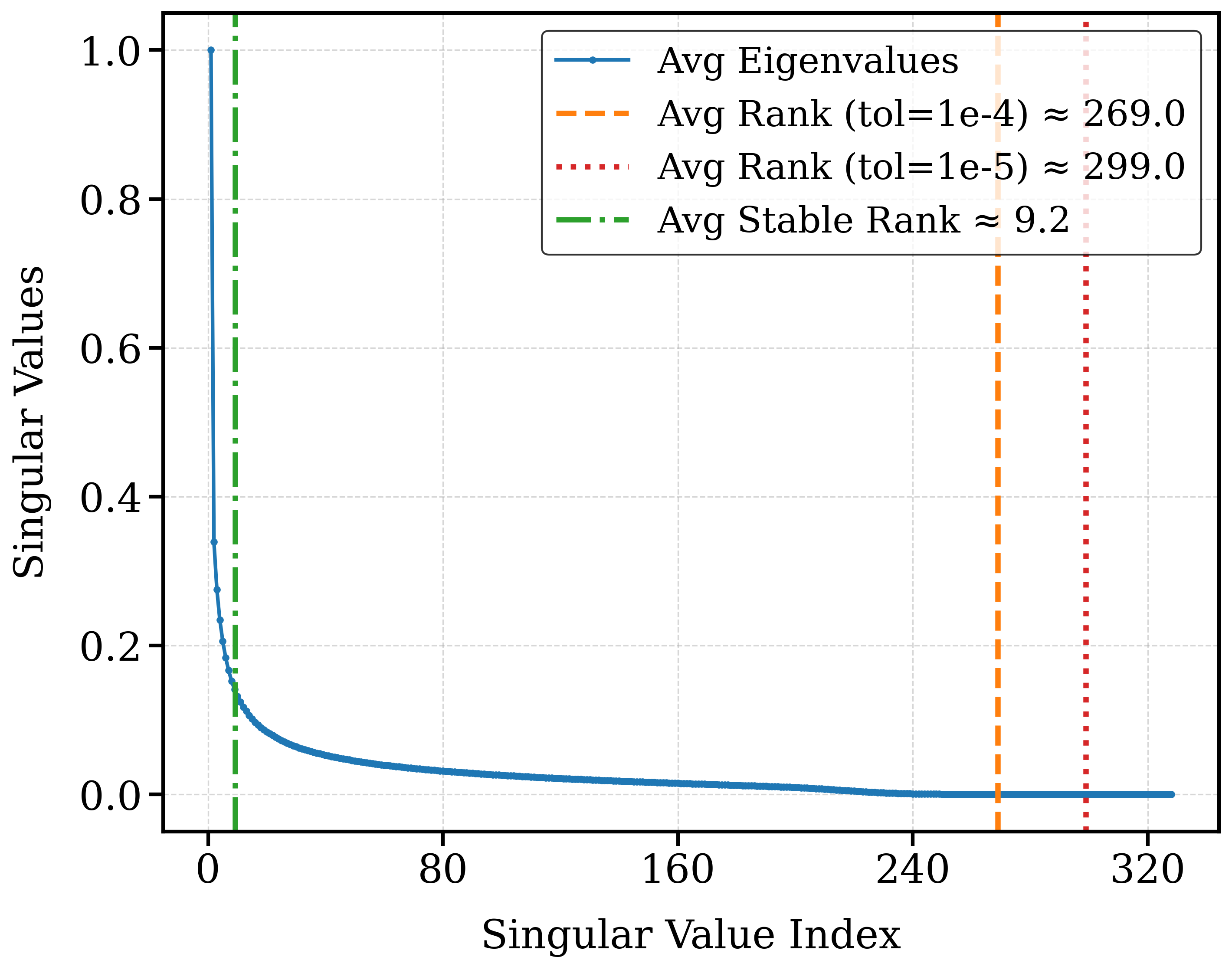}
  \caption{Average eigenvalue spectrum on HotpotQA using Llama-3-8B-Instruct model. It shows that a trivial change in the threshold leads to larger rank fluctuation, motivating the proposal of the stable rank.}
  \label{fig:avg_metrics}
\end{figure}

As illustrated in Figure \ref{fig:avg_metrics}, truncating at $1e-4$ yields a rank of 269, whereas a naive threshold at $1e-5$ results in a rank of 299, leading to a substantial difference in rank calculation. Both including substantial long-tail components. Therefore, we introduce the stable rank as a consistent and robust measure of effective dimensionality.

\begin{figure}
\begin{center}
    \begin{subfigure}[b]{0.32\textwidth}
        \centering
        \includegraphics[width=\textwidth]{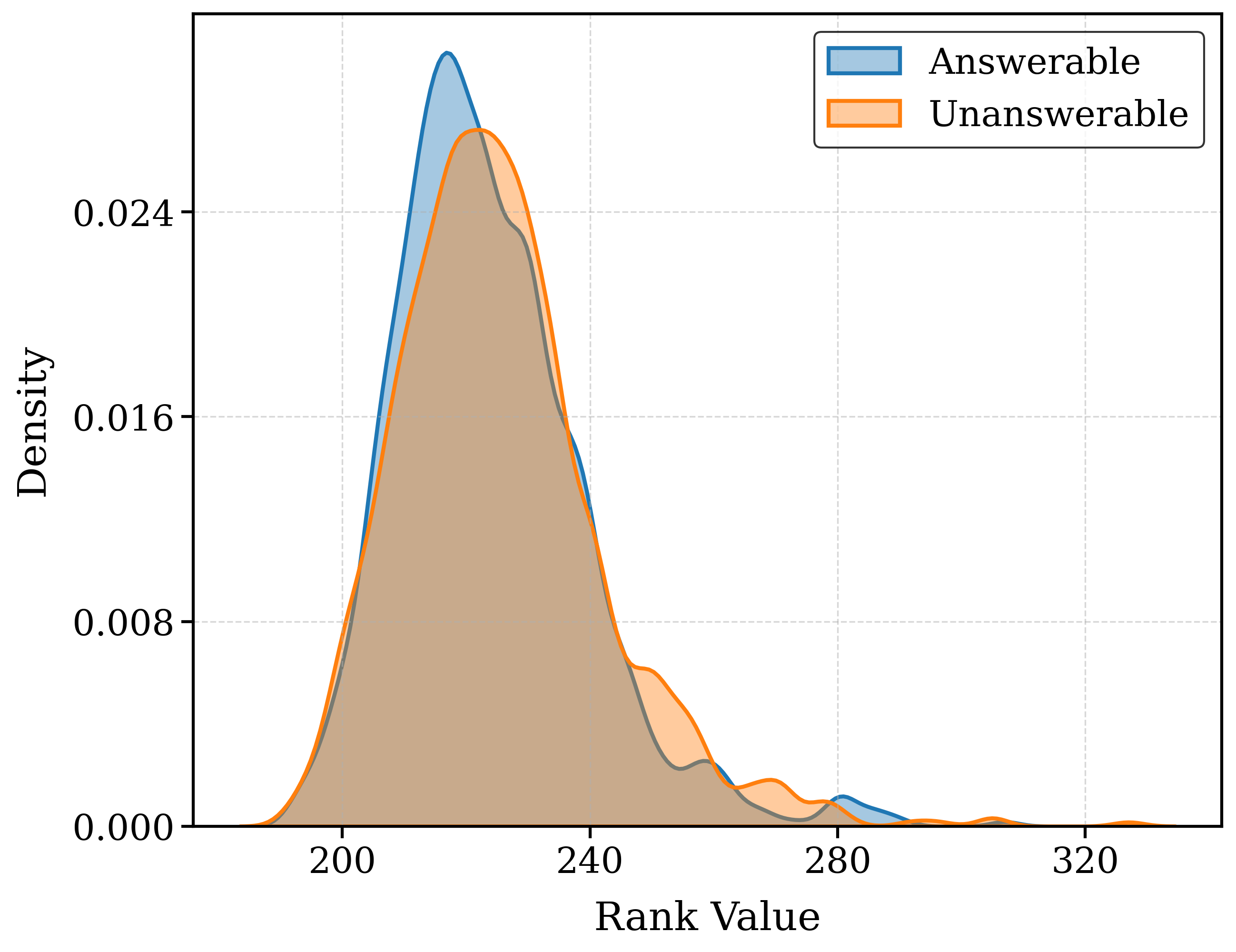}
        \caption{Hidden State Rank}
        \label{fig:hidden_state_rank}
    \end{subfigure}
    \begin{subfigure}[b]{0.32\textwidth}
        \centering
        \includegraphics[width=\textwidth]{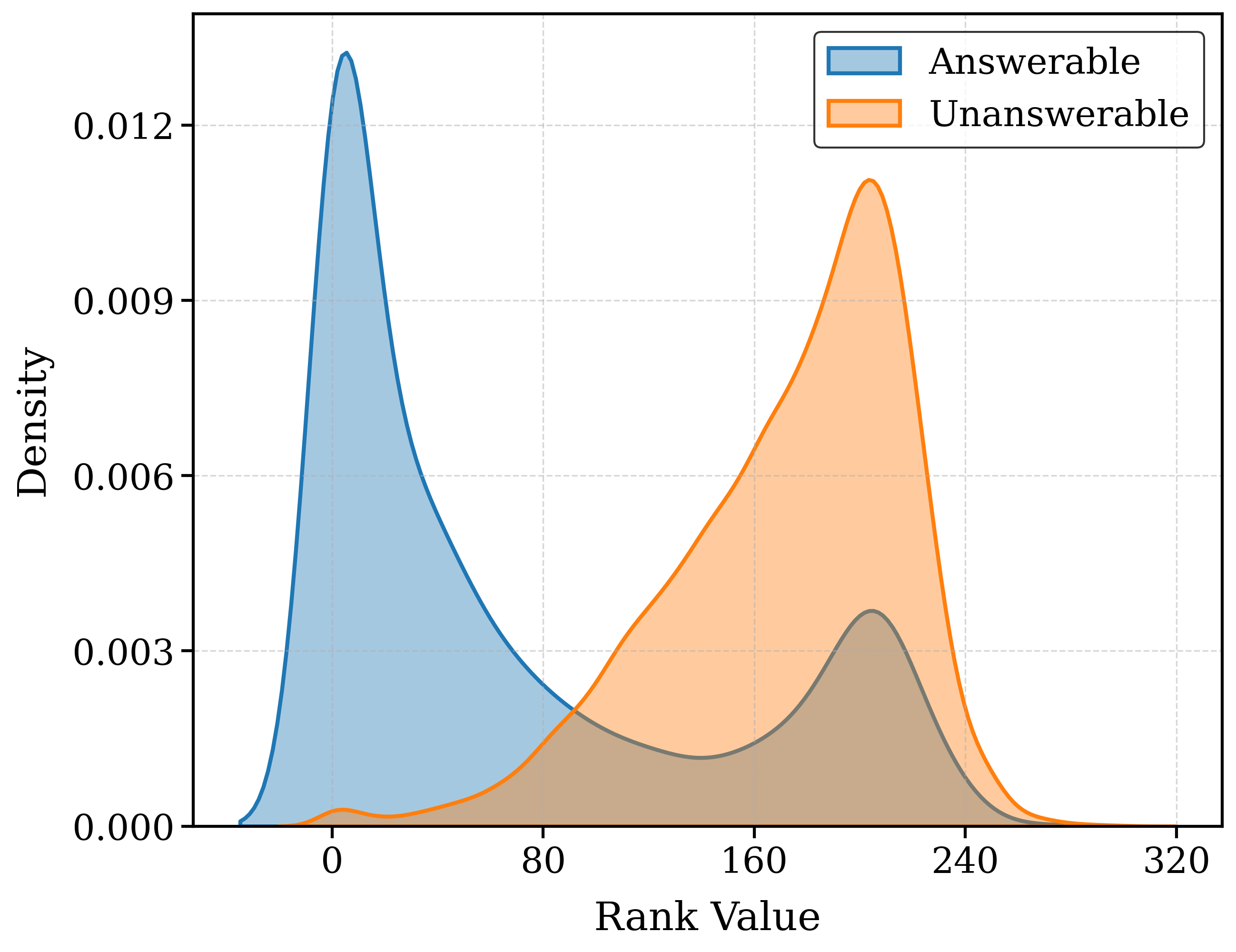}
        \caption{Gradient Rank}
        \label{fig:gradient_rank}
    \end{subfigure}
\end{center}
\caption{Density distributions of rank(Hidden States) in (a) and rank (Gradient) in (b) for Answerable and Unanswerable queries from the HotpotQA dataset. }
\label{fig:overall_rank_distribution}
\end{figure}

\begin{figure}[htbp]
    \centering
    \includegraphics[width=0.6\textwidth]{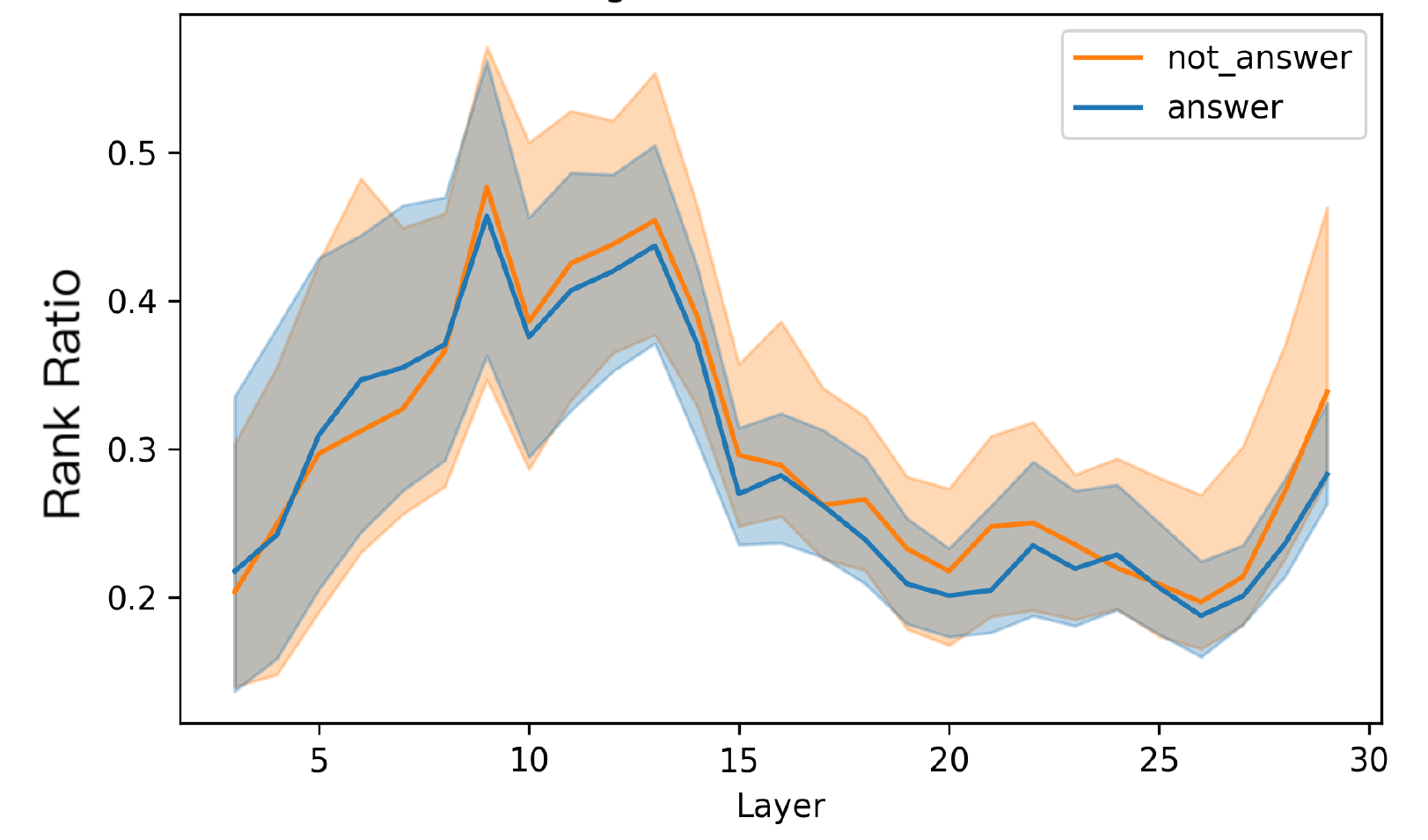} 
    \caption{Rank ratio across different layers for correctly and incorrectly answered samples.}
    \label{fig:layer_trend}
\end{figure}

\subsubsection{Effects of layer selections.} 
\begin{figure}[h]
    \centering
    \includegraphics[width=0.5\textwidth]{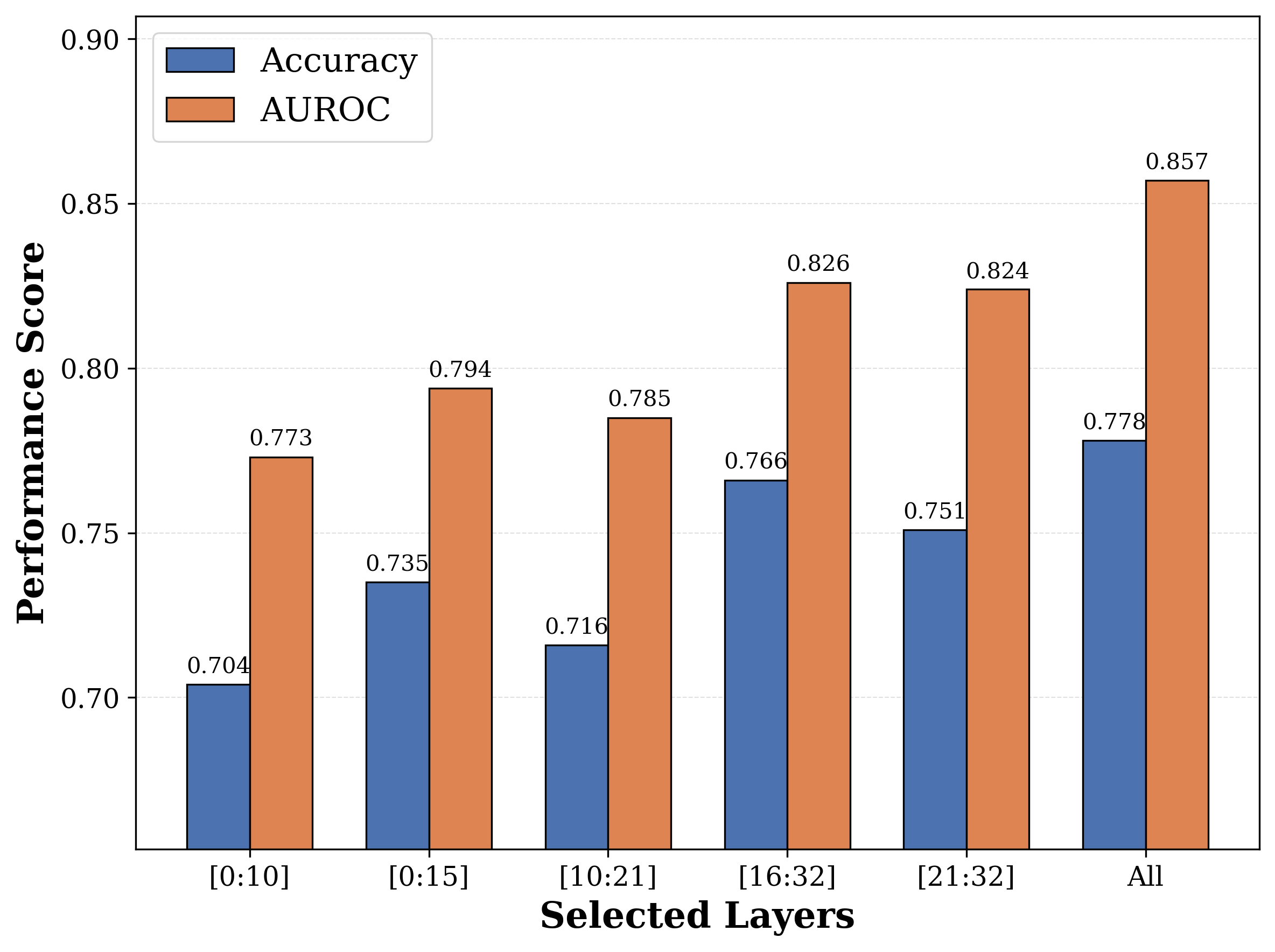}
    \caption{Performance comparison across different selected layers.}
    \label{fig:ablation_layer}
\end{figure}

\begin{itemize}[leftmargin=*, topsep=0pt]
    \item[] \textit{(a) Dynamics of the layer-wise rank ratio.} To elucidate why a holistic, cross-layer dynamic modelling is necessary, we show the trends for both answerable and unanswerable examples in Figure \ref{fig:layer_trend}. The distinguishability between answerable and unanswerable queries varies inconsistently across layers — no single layer dominates, and the absolute differences remain marginal throughout. This motivates a cross-layer approach that leverages all layer-wise representations to capture the nuanced dynamics underlying answerability.
 \item[] \textit{(b) Ablation of using rank ratios from different layers as probe input.} We conduct an ablation study on the choice of Transformer layers used as input features for the probe. Figure \ref{fig:ablation_layer} illustrates the classification performance when utilizing metrics from different layer subsets. The results demonstrate that aggregating signals from all layers yields the optimal performance, while utilizing later layers generally outperforms relying on earlier ones.
\end{itemize}


\subsubsection{Acc results of comparison between \modelname{} and the threshold based methods.} 
\begin{figure}[h]
    \centering
    \includegraphics[width=0.99\textwidth]{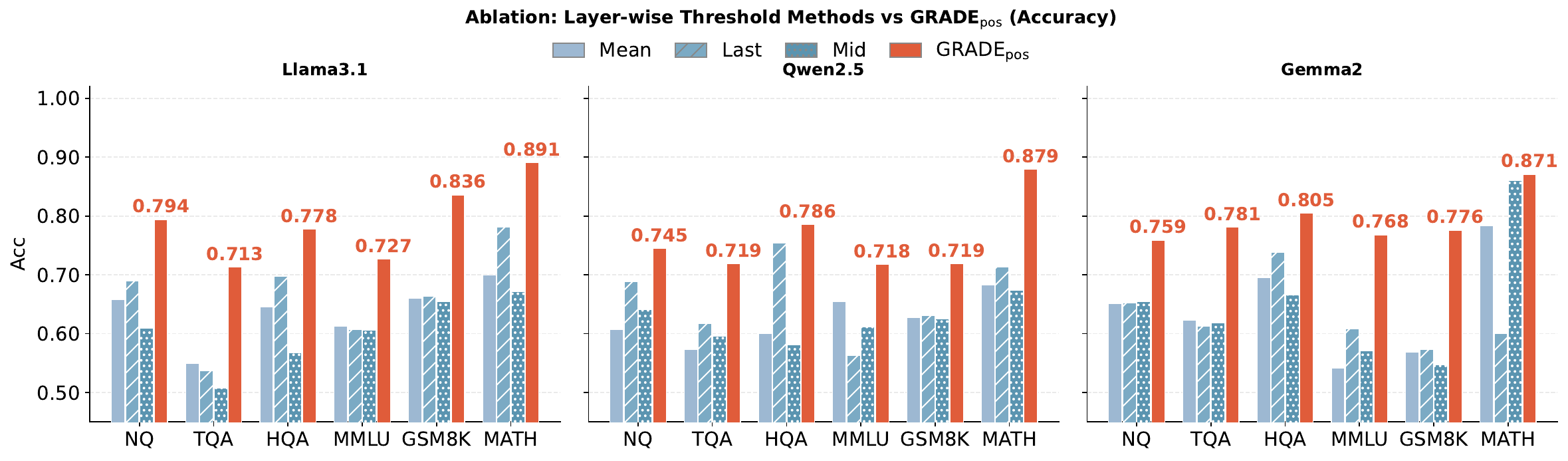}
    \caption{Comparison results (Acc) among detection threshold (the mean, last-layer, and middle-layer rank ratio, respectively) versus \modelname{} (rank ratio patterns cross different layers).}
    \label{fig:com_thres_acc}
\end{figure}

In addition to the AUROC results in Figure~\ref{tab:ablation_results}, we show the Acc in Figure~\ref{fig:com_thres_acc}. The trends are consistent that our \modelname{} performs better across all the datasets and models compared to the layer-wise threshold baselines (mean, last and mid shown here).

\subsubsection{Computational Complexity and Resource Overhead}
\label{supp:computational_overhead}

\modelname{} introduces minimal computational overhead, requiring only a single backward pass per reasoning step without parameter updates or iterative optimization. Table~\ref{tab:computational_cost} presents the additional computational time and memory deployment cost on GSM8K, along with their relative increase compared to the standard forward pass.

\begin{table}[h]
\centering
\small
\begin{tabular}{lcc}
\toprule
\textbf{Backbone} & \textbf{Extra Time} & \textbf{Extra Memory} \\
\midrule
Llama-8B & 0.063 s (2.2\%) & 2857.11 MB (17.1\%) \\
Qwen-7B  & 0.049 s (1.9\%) & 3266.42 MB (20.9\%) \\
Gemma-9B & 0.116 s (2.6\%) & 2527.18 MB (12.7\%) \\
\bottomrule
\end{tabular}
\caption{Extra computation time and memory deployment cost introduced by the backward pass per reasoning step on GSM8K, along with relative increase percentages compared to the forward pass.}
\label{tab:computational_cost}
\end{table}

Across all evaluated backbones, the backward pass introduces at most $0.116$ seconds (a maximum relative time increase of only $2.6\%$) per reasoning step. While computing gradients requires additional memory deployment for activations, the extra footprint remains bounded within $20.9\%$, demonstrating that \modelname{} is efficient and highly scalable for practical deployment.
\end{document}